\documentclass[10pt,journal,compsoc]{IEEEtran}

\ifCLASSOPTIONcompsoc
  \usepackage[nocompress]{cite}
\else
  \usepackage{cite}
\fi

\usepackage{graphicx}
\usepackage[english]{babel}
\usepackage{inputenc}
\usepackage[T1]{fontenc}
\usepackage{graphicx}
\usepackage{amssymb}
\usepackage{amsmath}
\usepackage{amsthm}
\usepackage{multirow}
\usepackage[usenames, dvipsnames]{xcolor}
\usepackage{epstopdf}
\usepackage{bm}
\usepackage{footnote}
\usepackage{stfloats}
\usepackage{placeins}
\usepackage{color,soul}
\usepackage{dsfont}
\usepackage{pbox}
\usepackage{enumerate}
\usepackage{etextools}
\usepackage{tabulary}
\usepackage{dsfont}
\usepackage{mathtools}
\mathtoolsset{showonlyrefs=true}

\usepackage{capt-of}
\usepackage{booktabs}
\usepackage{varwidth}
\usepackage{setspace}
\usepackage{algorithm}
\usepackage[noend]{algpseudocode}

\def\x{{\mathbf x}}
\def\X{{\mathbf X}}

\def\u{ \mathbf{u}}

\def\y{{\mathbf y}}

\def\z{{\mathbf z}}

\def\u{{\mathbf u}}

\def\D{{\mathbf D}}

\def\A{{\mathbf A}}
\def\B{{\mathbf B}}
\def\I{{\mathbf I}}

\def\Q{{\mathbf Q}}
\def\0{{\mathbf 0}}
\def\W{{\mathbf W}}
\def\w{{\mathbf w}}

\def\b{{\mathbf b}}

\def\b{{\mathbf b}}

\def\alfa{{\boldsymbol \alpha}}
\def\gama{{\boldsymbol \gamma}}
\def\gamat{\tilde{\boldsymbol \gamma}}

\def\prox{{\text{prox}}}

\newtheorem{theorem}{Theorem}[section]
\newtheorem{corollary}{Corollary}[theorem]
\newtheorem{lemma}[theorem]{Lemma}

\begin{document}

\newcommand{\notej}[1]{\textcolor{black}{#1}}

\title{On Multi-Layer Basis Pursuit, Efficient Algorithms and Convolutional Neural Networks}

\author{Jeremias~Sulam,~\IEEEmembership{Member,~IEEE,}
	Aviad~Aberdam,
	Amir~Beck,
	Michael~Elad,~\IEEEmembership{Fellow,~IEEE}
	\IEEEcompsocitemizethanks{\IEEEcompsocthanksitem J. Sulam and A. Aberdam contributed equally to this work.}
	\IEEEcompsocitemizethanks{\IEEEcompsocthanksitem J. Sulam is with the Department of Biomedical Engineering, Johns Hopkins University. A. Aberdam is with the Department of Electrical Engineering, Technion -- Israel Institute of Technology. A. Beck is with the School of Mathematical Sciences at Tel Aviv University. M. Elad is with the Department of Computer Science, Technion -- Israel Institute of Technology.}
	\thanks{The research leading to these results has received funding from the European Research Council under European Unions Seventh Framework Programme, ERC Grant agreement no. 320649. The work of Amir Beck was partially supported by the Israel Science Foundation 1821/16.}
}

\markboth{IEEE Transactions on Pattern Analysis and Machine Intelligence, VOL., No., November~2018}%
{Shell \MakeLowercase{\textit{et al.}}: Bare Demo of IEEEtran.cls for Computer Society Journals}

\IEEEtitleabstractindextext{%
	\begin{abstract}

Parsimonious representations are ubiquitous in modeling and processing information. Motivated by the recent Multi-Layer Convolutional Sparse Coding (ML-CSC) model, we herein generalize the traditional Basis Pursuit problem to a multi-layer setting, introducing similar sparse enforcing penalties at different representation layers in a symbiotic relation between synthesis and analysis sparse priors. We explore different iterative methods to solve this new problem in practice, and we propose a new Multi-Layer Iterative Soft Thresholding Algorithm (ML-ISTA), as well as a fast version (ML-FISTA). We show that these nested first order algorithms converge, in the sense that the function value of near-fixed points can get arbitrarily close to the solution of the original problem. 

We further show how these algorithms effectively implement particular recurrent convolutional neural networks (CNNs) that generalize feed-forward ones without introducing any parameters. We present and analyze different architectures resulting unfolding the iterations of the proposed pursuit algorithms, including a new Learned ML-ISTA, providing a principled way to construct deep recurrent CNNs. Unlike other similar constructions, these architectures unfold a global pursuit holistically for the entire network. We demonstrate the emerging constructions in a supervised learning setting, consistently improving the performance of classical CNNs while maintaining the number of parameters constant.

	\end{abstract}

	}

\maketitle


\section{Introduction}

Sparsity has been shown to be a driving force in a myriad of applications in computer vision \cite{wright2010sparse,zeiler2010deconvolutional,mairal2010online}, statistics \cite{tibshirani2011regression,tibshirani2015statistical} and machine learning \cite{kavukcuoglu2010fast,henaff2011unsupervised,huang2007unsupervised}. Most often, sparsity is often enforced not on a particular signal but rather on its representation in a transform domain. Formally, a signal $\x \in \mathbb{R}^n$ admits a sparse representation in terms of a dictionary $\D\in \mathbb{R}^{n\times m}$ if $\x = \D\gama$, and $\gama\in \mathbb{R}^m$ is sparse. In its simplest form, the problem of seeking for a sparse representation for a signal, possibly contaminated with noise $\w$ as $\y = \x + \w$, can be posed in terms of the following pursuit problem:
\begin{equation} \label{eq:P0}
\min_{\gama}\ \| \gama \|_0 \ \text{ s.t. }  \|\y - \D\gama \|_2^2 \leq \varepsilon,
\end{equation}
where the $\ell_0$ pseudo-norm counts the number of non-zero elements in $\gama$. The choice of the (typically overcomplete) dictionary $\D$ is far from trivial, and has motivated the design of different analytic transforms \cite{candes2000curvelets,do2005contourlet,kutyniok2012shearlets} and the development of dictionary learning methods \cite{aharon2006ksvd,sulam2016trainlets,mairal2010online}. The above problem, which is  NP-hard in general, is often relaxed by employing the $\ell_1$ penalty as a surrogate for the non-convex $\ell_0$ measure, resulting in the celebrated Basis Pursuit De-Noising (BPDN) problem\footnote{This problem is also known in the statistical learning community as Least Absolute Shrinkage and Selection Operator (LASSO) \cite{tibshirani2011regression}, where the matrix $\D$ is given by a set of measurements or descriptors, in the context of a sparse regression problem.}:
\begin{equation} \label{eq:Lasso}
\min_{\gama}\ \lambda \| \gama \|_1 + \frac{1}{2}  \|\y - \D\gama \|_2^2.
\end{equation}
The transition from the $\ell_0$ to the relaxed $\ell_1$ case is by now well understood, and the solutions to both problems do coincide under sparse assumptions on the underlying representation (in the noiseless case), or have shown to be close enough in more general settings \cite{donoho2003optimally,tropp2006just}.

This traditional model was recently extended to a multi-layer setting \cite{papyan2016convolutional,Aberdam2018Holistic}, where a signal is expressed as $\x  = \D_1\gama_1$, for a sparse $\gama_1\in \mathbb{R}^{m_1}$ and (possibly convolutional) matrix $\D_1$, while also assuming that this representation satisfies $\gama_1 = \D_2\gama_2$, for yet another dictionary $\D_2\in \mathbb{R}^{m_1\times m_2}$ and sparse $\gama_2\in \mathbb{R}^{m_2}$. Such a construction can be cascaded for a number of $L$ layers\footnote{In the convolutional setting \cite{papyan2017working,sulam2017multi}, the notion of sparsity is better characterized by the $\ell_{0,\infty}$ pseudo-norm, which quantifies the \emph{density of non-zeros} in the convolutional representations in a local sense. Importantly, however, the BPDN formulation (i.e., employing an $\ell_1$ penalty), still serves as a proxy for this $\ell_{0,\infty}$ norm. We refer the reader to \cite{papyan2017working} for a thorough analysis of convolutional sparse representations.}. Under this framework, given the measurement $\y$, this multi-layer pursuit problem (or Deep Coding Problem, as first coined in \cite{papyan2016convolutional}), can 
be expressed as
\begin{equation} \label{eq:DCP}
\begin{split}
\underset{\{\gama_i\}}{\min} ~ \|\y- \D_1\gama_1\|^2_2 \quad  \text{s.t.} \ \left\{ \gama_{i-1} = \D_i \gama_i, ~ \|\gama_i\|_{0} \leq s_i \right\}_{i=1}^L,
\end{split}
\end{equation}
with $\x = \gama_0$.
In this manner, one searches for the closest signal to $\y$ while satisfying the model assumptions. This can be understood and analyzed as a projection problem \cite{sulam2017multi}, providing an estimate such that $\hat{\x} = \D_1\hat{\gama}_1 = \D_1\D_2\hat{\gama}_2 = \dots = \D_{(1,L)}\hat{\gama}_L$, while forcing all intermediate representations to be sparse. Note the notation $\D_{(i,L)} = \D_i\dots\D_L$ for brevity. 
Remarkably, the forward pass of neural networks (whose weights at each layer, $\W_i$, are set as the transpose of each dictionary $\D_i$) yields stable estimations for the intermediate features or representations $\hat{\gama}_i$ provided these are sparse enough \cite{papyan2016convolutional}. \notej{Other generative models have also been recently proposed (as the closely related probabilistic framework in \cite{patel2016probabilistic,nguyen2016semi}), but the multi-layer sparse model provides a convenient way to study deep learning architectures in terms of pursuit algorithms \cite{Papyan18}.}

As an alternative to the forward pass, one can address the problem in Equation \eqref{eq:DCP} by adopting a projection interpretation and develop an algorithm based on a global pursuit, as in \cite{sulam2017multi}. More recently, the work in \cite{Aberdam2018Holistic} showed that this problem can be cast as imposing an analysis prior on the signal's deepest sparse representation. Indeed, the problem in \eqref{eq:DCP} can be written concisely as:
\begin{align} \label{eq:DCP_3}
& \underset{\{\gama_i\}}{\min} \quad \|\y- \D_{(1,L)}\gama_L\|^2_2 \\ & \text{s.t.} \quad \|\gama_L\|_0 \leq s_L, ~ \left\{ \|\D_{(i,L)}\gama_L\|_0 \leq s_{i-1} \right\}^L_{i=1}.
\end{align}
This formulation explicitly shows that the intermediate dictionaries $\D_{(i,L)}$ play the role of analysis operators, resulting in a representation $\gama_L$ which should be orthogonal to as many rows from $\D_{(i,L)}$ as possible --  so as to produce zeros in $\gama_i$. Interestingly, this analysis also allows for \emph{less sparse} representations in shallower layers while still being consistent with the multi-layer sparse model. 
While a pursuit algorithm addressing \eqref{eq:DCP_3} was presented in \cite{Aberdam2018Holistic}, it is greedy in nature and does not scale to high dimensional signals. In other words, there are currently no efficient pursuit algorithms for signals in this multi-layer model that leverage this symbiotic analysis-synthesis priors. More importantly, it is still unclear how the dictionaries could be trained from real data under this scheme. \notej{These questions are fundamental if one is to bridge the theoretical benefits of the multi-layer sparse model with practical deep learning algorithms.}

In this work we propose a relaxation of the problem in Equation \eqref{eq:DCP_3}, turning this seemingly complex pursuit into a convex multi-layer generalization of the Basis Pursuit (BP) problem\footnote{In an abuse of terminology, and for the sake of simplicity, we will refer to the BPDN problem in Equation \eqref{eq:Lasso} as BP.}. Such a formulation, to the best of our knowledge, has never before been proposed nor studied, though we will comment on a few particular and related cases that have been of interest to the image processing and compressed sensing communities. We explore different algorithms to solve this multi-layer problem, such as variable splitting and the Alternating Directions Method of Multipliers (ADMM) \cite{B83,boyd2011distributed} and the Smooth-FISTA from \cite{beck2012smoothing}, and we will present and analyze two new generalizations of Iterative Soft Thresholding Algorithms (ISTA).
We will further show that these algorithms generalize feed-forward neural networks (NNs), both fully-connected and convolutional (CNNs), in a natural way. More precisely: the first iteration of such algorithms implements a traditional CNN, while a new recurrent architecture emerges with subsequent iterations. In this manner, the proposed algorithms provide a principled framework for the design of recurrent architectures. While other works have indeed explored the unrolling of iterative algorithms in terms of CNNs (e.g. \cite{zhang2018ista,murdock2018deep}), we are not aware of any work that has attempted nor studied the unrolling of a global pursuit with convergence guarantees. Lastly, we demonstrate the performance of these networks in practice by training our models for image classification, consistently improving on the classical feed-forward architectures without introducing filters nor any other extra parameters in the model.


\section{Multi-Layer Basis Pursuit}

In this work we propose a convex relaxation of the problem in Equation \eqref{eq:DCP_3}, resulting in a multi-layer BP problem. For the sake of clarity, we will limit our formulations to two layers, but these can be naturally extended to multiple layers -- as we will effectively do in the experimental section. This work is centered around the following problem:
\begin{equation} \label{eq:MLlasso}
(P): \quad \min_{\gama} \frac{1}{2} \|\y - \D_1\D_2\gama \|_2^2 + \lambda_1 \|\D_2\gama \|_1 + \lambda_2\| \gama \|_1.
\end{equation}

This formulation imposes a particular mixture of synthesis and analysis priors. Indeed, if $\lambda_2$>0 and $\lambda_1 = 0$, one recovers a traditional Basis Pursuit formulation with a factorized global dictionary. If $\lambda_1>0$, however, an analysis prior is enforced on the representation $\gama$ by means of $\D_2$, resulting in a more regularized solution. Note that if $\lambda_2 = 0$, $\lambda_1>0$ and $\ker\D_2$ is not empty, the problem above becomes ill-posed without a unique solution\footnote{It is true that also in Basis Pursuit one can potentially obtain infinite solutions, as the problem is not strongly convex.} since $\ker\D_{(1,2)} \cap \ker\D_2 \neq \{ \mathbf{0} \}$. In addition,
unlike previous interpretations of the multi-layer sparse model (\cite{papyan2016convolutional,sulam2017multi,Aberdam2018Holistic}), our formulation stresses the fact that there is \emph{one unknown} variable, $\gama$, with different  priors enforced on it. Clearly, one may also define and introduce $\gama_1 = \D_2\gama$, but this should be interpreted merely as the introduction of auxiliary variables to aid the derivation and interpretation of the respective algorithms. We will expand on this point in later sections.

Other optimization problems similar to $(P)$ have indeed been proposed, such as the Analysis-LASSO \cite{candes2010compressed,lin2014sparse}, though their observation matrix and the analysis operator ($\D_{(1,2)}$ and $\D_2$, in our case) must be independent, and the latter is further required to be a tight frame \cite{candes2010compressed,lin2014sparse}. The Generalized Lasso problem \cite{tibshirani2011solution} is also related to our multi-layer BP formulation, as we will see in the following section. On the other hand, and in the context of image restoration, the work in \cite{bioucas2008iterative} imposes a Total Variation and a sparse prior on the unknown image, as does the work in \cite{haeffele14}, thus being closely related to the general expression in \eqref{eq:MLlasso}. 


\subsection{Algorithms}

From an optimization perspective, our multi-layer BP problem can be expressed more generally as 
\begin{equation} \label{eq:MLlasso_general}
\min_{\gama}\ F(\gama) =   f(\D_2\gama) + g_1(\D_2\gama) + g_2(\gama),
\end{equation}
where $f$ is convex and smooth, and $g_1$ and $g_2$ are convex but non-smooth. For the specific problem in \eqref{eq:MLlasso}, $f(\z) = \frac{1}{2}\|\y - \D_1\z\|^2_2$, $g_1(\z) = \lambda_1\|\z\|_1$ and $g_2(\z) =\lambda_2\|\z\|_1$.
Since this problem is convex, the choice of available algorithms is extensive. We are interested in high-dimensional settings, however, where interior-point methods and other solvers depending on second-order information might have a prohibitive computational complexity. In this context, the Iterative Soft Thresholding Algorithm (ISTA), and its Fast version (FISTA), are appealing as they only require matrix-vector multiplications and entry-wise operations. The former, originally introduced in \cite{daubechies2004iterative}, provides convergence (in function value) of order $\mathcal{O}(1/k)$, while the latter provides an improved convergence rate with order of $\mathcal{O}(1/k^2)$ \cite{beck2009fast}.

Iterative shrinkage algorithms decompose the total loss into two terms: $f(\gama)$, convex and smooth (with Lipschitz constant $L$), and $g(\gama)$, convex and possibly non smooth. The central idea of ISTA, as a proximal gradient method for finding a minimizer of $f + g$, is to iterate the updates given by the proximal operator of $g(\cdot)$ at the forward-step point:
\begin{equation} \label{eq:ista_step}
\gama^{k+1} = \prox_{\frac{1}{L}g}\left( \gama^k - \frac{1}{L} \nabla f(\gama^k) \right).
\end{equation}
Clearly, the appeal of ISTA depends on how effectively the proximal operator can be computed. When $g(\gama) = \lambda \|\gama\|_1$ (as in the original BP formulation), such a proximal mapping becomes separable, resulting in the element-wise shrinkage or soft-thresholding operator. However, this family of methods cannot be readily applied to $(P)$ where $g(\gama) =  \lambda_1 \|\D_2\gama \|_1 + \lambda_2\| \gama \|_1$. Indeed, computing $\prox_{g}(\gama)$ when $g(\cdot)$ is a sum of $\ell_1$ \emph{composite} terms is no longer directly separable, and one must resort to iterative approaches, making ISTA lose its appeal. 

The problem $(P)$ is also related to the Generalized Lasso formulation \cite{tibshirani2011solution} in the compressed sensing community, which reads
\begin{equation}\label{eq:generalizedLasso}
\min_{\gama} \frac{1}{2} \|\y - \X \gama \|_2^2 + \nu \|\A\gama \|_1.
\end{equation}
Certainly, the multi-layer BP problem we study can be seen as a particular case of this formulation\footnote{One can rewrite problem \eqref{eq:MLlasso} as in \eqref{eq:generalizedLasso} by making $\X = \D_{(1,L)}$ and $\A = [\lambda_1/\nu \D_2^T, \lambda_2/\nu \mathbf{I}]^T$.}. With this insight, one might consider solving $(P)$ through the solution of the generalized Lasso \cite{tibshirani2011solution}. However, such an approach also becomes computationally demanding as it boils down to an iterative algorithm that includes the inversion of linear operators. Other possible solvers might rely on re-weighted $\ell_2$ approaches \cite{chartrand2008iterative}, but these also require iterative matrix inversions.

\begin{algorithm}[t]
 \caption{ADMM algorithm for a two-layer ML-CSC model.}
 \label{alg:ADMM}
Input: signal $\y$, dictionaries $\D_i$ and parameters $\lambda_i$
\begin{algorithmic}[1]
 \While{not converged} \\
 \hspace*{\algorithmicindent} 
\begin{minipage}{\textwidth}
	$\gama_2 \gets  \arg\min_{\gama_2} \| \y - \D_1\D_2\gama_2 \|_2^2$ \\
	$ \phantom{aaaaaa} + \frac{\rho}{2} \| \gama_1 - \D_2\gama_2 + \mathbf{u}\|_2^2 + \lambda_2 \|\gama_2\|_1 $
\end{minipage} \\ 
\hspace*{\algorithmicindent} $\gama_1 \gets  \arg\min_{\gama_1} \frac{\rho}{2} \| \gama_1 - \D_2\gama_2 + \mathbf{u}\|_2^2 + \lambda_1 \|\gama_1\|_1$ \\ \vspace{.1cm}
\hspace*{\algorithmicindent} $\mathbf{u}\gets \mathbf{u}+ \rho (  \gama_1 - \D_2\gama_2  ) $ 
 \EndWhile
\end{algorithmic}
\end{algorithm}

A simple way of tackling problem $(P)$ is the popular Alternating Directions Method of Multipliers (ADMM), which provides a natural way to address these kind of problems through variable splitting and auxiliary variables. For a two layer model, one can rewrite the multi-layer BP as a constrained minimization problem:
\begin{align} \label{eq:Split_constraint}
& \min_{\gama_1,\gama_2} ~ \frac{1}{2} \|\y - \D_1\D_2\gama_2 \|_2^2 + \lambda_1 \|\gama_1 \|_1 + \lambda_2\| \gama_2 \|_1  \\ & \text{ s.t. }\quad \gama_1 = \D_2\gama_2.
\end{align}
ADMM minimizes this constrained loss by constructing an augmented Lagrangian (in normalized form) as
\begin{multline} \label{eq:Split_constraint}
\min_{\gama_1,\gama_2,\u} ~ \frac{1}{2} \|\y - \D_1\D_2\gama_2 \|_2^2 + \lambda_1 \|\gama_1 \|_1 \\ + \lambda_2\| \gama_2 \|_1  + \frac{\rho}{2} \|  \gama_1 - \D_2\gama_2 + \mathbf{u}\|_2^2,
\end{multline}
which can be minimized iteratively by repeating the updates in Algorithm \ref{alg:ADMM}. This way, and after merging both $\ell_2$ terms, the pursuit of the inner-most representation ($\gama_2$ in this case) is carried out in terms of a regular BP formulation that can be tackled with a variety of convex methods, including ISTA or FISTA. The algorithm then updates the intermediate representations ($\gama_1$) by a simple shrinkage operation, followed by the update of the dual variable, $\u$. Note that this algorithm is guaranteed to converge (at least in the sequence sense) to a global optimum of $(P)$ due to the convexity of the function being minimized  \cite{B83,boyd2011distributed}. 


A third alternative, which does not incur in an additional inner iteration nor inversions, is the Smooth-FISTA approach from \cite{beck2012smoothing}. S-FISTA addresses cost functions of the same form as problem $(P)$ by replacing one of the non-smooth functions, $g_1(\gama)$ in Eq. \eqref{eq:MLlasso_general}, by a smoothed version in terms of its Moreau envelope. In this way, S-FISTA converges with order $\mathcal{O}(1/\varepsilon)$ to an estimate that is $\varepsilon$-away from the solution of the original problem in terms of function value. We will revisit this method further in the following sections.

Before moving on, we make a short intermission to note here that other Basis Pursuit schemes have been proposed in the context of multi-layer sparse models and deep learning. Already in \cite{papyan2016convolutional} the authors proposed the Layered Basis Pursuit, which addresses the sequence of pursuits given by
\begin{equation}
\hat{\gama}_i \gets \underset{\gama_i}{\arg\min} ~ \| \hat{\gama}_{i-1} - \D_i \gama_i \|^2_2 + \lambda_i \|\gama_i\|_1,
\end{equation}
from $i=1$ to $L$, where $\gama_0 = \y$. Clearly, each of these can be solved with any BP solver just as well. \notej{A related idea was also recently proposed in \cite{sun2018supervised}, showing that cascading basis pursuit problems can lead to competitive deep learning constructions}. However, the Layered Basis Pursuit formulation, or other similar variations that attempt to \emph{unfold} neural network architectures \cite{murdock2018deep,zhang2018ista}, do not minimize $(P)$ and thus their solutions only represent sub-optimal and heuristic approximations to the minimizer of the multi-layer BP. More clearly, such a series of steps never provide estimates $\hat{\gama}_i$ that can generate a signal according to the multi-layer sparse model. As a result, one cannot reconstruct $\hat{\x} = \D_{(1,L)}\hat{\gama}_L$, because each representation is required to explain the next layer \emph{only approximately}, so that $\hat{\gama}_{i-1}\neq\D_i\hat{\gama}_i$.

\subsection{Towards Multi-Layer ISTA}

We now move to derive the proposed approach to efficiently tackle $(P)$ while relying on the concept of the \emph{gradient mapping} (see for example \cite[Chapter 10]{B17}), which we briefly review next. Given a function $F(\gama) = f(\gama) + g(\gama)$, where $f$ is convex and smooth with Lipschitz constant $L$ and $g$ is convex, the gradient mapping is the operator given by
\begin{equation}
    G_L^{f,g}(\gama) = L\left[ \gama - \prox_{\frac{1}{L}g}\left(\gama - \frac{1}{L}\nabla f(\gama) \right) \right].
\end{equation}
Naturally, the ISTA update step in Equation \eqref{eq:ista_step} can be seen as a ``gradient-mapping descent'' step, since it can be rewritten as $\gama^{k+1} = \gama^k - \frac{1}{L} G_L^{f,g}(\gama)$. Moreover, $G_L^{f,g}(\gama)$ provides a sort of generalization of the gradient of $F(\gama)$, since
\begin{enumerate}
    \item $ G_L^{f,g}(\gama) = \nabla F(\gama) = \nabla f(\gama)$ if $g(\gama)\equiv0$,
    \item $G_L^{f,g}(\gama)=0$ if and only if $\gama$ is a minimizer of $F(\gama)$.
\end{enumerate}
We refer the reader to \cite[Chapter 10]{B17} for further details on gradient mapping operators.
 
Returning to the problem in \eqref{eq:MLlasso_general}, our first attempt to minimize $F(\gama_2) =   f(\D_2\gama_2) + g_1(\D_2\gama_2) + g_2(\gama_2)$ is a \emph{proximal gradient-mapping} method, and it takes an update of the following form:
\begin{equation} \label{eq:gofista}
\gama_2^{k+1} = \prox_{tg_2} \left( \gama_2^k - t\ G_{1/\mu}^{f(\cdot),g_1(\D_2\cdot)}(\gama_2^k) \right),
\end{equation}
for constants $\mu>0$ and $t>0$ that will be specified shortly. This expression, however, requires the computation of $\prox_{g_1(\D_2\cdot)}(\cdot)$, which is problematic as it involves a composite term\footnote{The proximal of a composition with an affine map is only available for unitary linear transformations. See \cite[Chapter 10]{B17} and \cite{combettes2011proximal} for further details.}. To circumvent this difficulty, we propose the following approximation in terms of $\gama_1 = \D_2\gama_2$. In the spirit of the chain rule\footnote{The step taken to arrive at Equation \eqref{eq:ml_ista} is not actually the chain rule, as the gradient mapping $G_L^{f,g}$ is not necessarily a gradient of a smooth function.}, we modify the previous update to
\begin{equation} \label{eq:ml_ista}
\gama_2^{k+1} = \prox_{tg_2} \left( \gama_2^k - t\ \D_2^T\  G_{1/\mu}^{f,g_1}(\gama^k_1) \right).
\end{equation}
Importantly, the above update step now involves the $\prox$ of $g_1(\cdot)$ as opposed to that of $g_1(\D_2\cdot)$. 
This way, in the case of $g_1(\cdot) = \lambda_1 \|\cdot\|_1$, the proximal mapping of $g_1$ becomes the soft-thresholding operator with parameter $\lambda_1$, i.e. $\prox_{g_1}(\gama_1) = \mathcal{T}_{\lambda_1} (\gama_1)$. An analogous operator is obtained for $\prox_{g_2}$ just as well. Therefore, the proposed Multi-Layer ISTA update can be concisely written as
\begin{equation} \label{eq:MLISTA_update}
\resizebox{1\hsize}{!}{$\gama_2^{k+1} = \mathcal{T}_{t\lambda_2} \left( \gama_2^k - \frac{t}{\mu} \D_2^T \left(\gama_1^k - \mathcal{T}_{\mu\lambda_1}(\gama^k_1 - \mu \D^T_1 (\D_1\gama^k_1 - \y))\right)\right)$}.
\end{equation}

A few comments are in place. First, this algorithm results in a nested series of shrinkage operators, involving only matrix-vector multiplications and entry-wise non linear operations. Note that if $\lambda_1 = 0$, i.e. in the case of a traditional Basis Pursuit problem, the update above reduces to the update of ISTA. Second, though seemingly complicated at first sight, the resulting operator in \eqref{eq:MLISTA_update} can be decomposed into simple recursive layer-wise operations, as presented in Algorithm \ref{alg:ML-ISTA}. Lastly, because the above update provides a multi-layer extension to ISTA, one can naturally suggest a ``fast version'' of it by including a momentum term, just as done by FISTA. In other words, ML-FISTA will be given by the iterations
\begin{align} \label{eq:ml-fista-update}
    \gama_2^{k+1} &= \prox_{tg_2} \left( \z^k - t \D_2^T G^{f,g_1}_{1/\mu}(\D_2\z^k) \right), \\
    \z^{k+1} &= \gama^{k+1}_2 + \rho^k (\gama_2^{k+1} - \gama_2^{k}),
\end{align}
where $\rho^k = \frac{t_k - 1}{t_{k+1}}$, and the $t_k$ parameter is updated according to $t_{k+1} = \frac{1+\sqrt{1+4t_k^2}}{2}$. Clearly, one can also write this algorithm in terms of layer-wise operations, as described in Algorithm \ref{alg:ML-FISTA}.

\begin{algorithm} \setstretch{1.3}
 \caption{Multi-Layer ISTA.}
 \label{alg:ML-ISTA}
 Input: {signal $\y$, dictionaries $\D_i$ and parameters $\lambda_i$.} \\
 Init: Set $\gama^k_0 = \y ~\forall~k$ and $\gama_L^1 = 0$.
\begin{algorithmic}[1] \small
 \For{$k = 1 : K$} \hspace{.9cm} \texttt{\color{gray}\% for each iteration} \\ 
 \hspace{.35cm} $\hat{\gama}_{i} \gets \D_{(i,L)}\gama^k_{L} \quad \ \forall i \in [0,L-1]$
\For{i = 1 : L} \hspace{.9cm} \texttt{\color{gray} \% for each layer} \\
\hspace{.7cm}$~~\gama^{k+1}_i \gets \mathcal{T}_{\mu_i\lambda_i} \left( \hat{\gama}_i - \mu_i\D^T_i (\D_i \hat{\gama}_i - \gama^{k+1}_{i-1} ) \right)$
 \EndFor
 \EndFor
 \end{algorithmic}
\end{algorithm}

\begin{algorithm} \setstretch{1.3}
 \caption{Multi-Layer FISTA.}
 \label{alg:ML-FISTA}
  Input: {signal $\y$, dictionaries $\D_i$ and parameters $\lambda_i$.} \\
 Set $\gama^k_0 = \y ~\forall~k$ and $\z = 0$.
\begin{algorithmic}[1] \small
 \For{$k = 1 : K$} \hspace{.9cm} \texttt{\color{gray}\% for each iteration} \\
\hspace{.4cm}$\hat{\gama}_{i} \gets \D_{(i,L)}\z \quad \forall i \in [0,L-1]$ 
\For{i = 1 : L} \hspace{.9cm} \texttt{\color{gray}\% for each layer}\\
\hspace{.2cm}$\qquad \gama^{k+1}_i \gets \mathcal{T}_{\mu_i\lambda_i} \left( \hat{\gama}_i - \mu_i \D^T_i (\D_i \hat{\gama}_i - \gama^{k+1}_{i-1} ) \right)$
 \EndFor \\
\hspace{.4cm} $t_{k+1} \gets \frac{1+\sqrt{1+4t_k^2}}{2}$ \\
\hspace{.4cm} $\z \gets \gama^{k+1}_L + \frac{t_k-1}{t_{k+1}} (\gama_L^{k+1} - \gama_L^{k})$
 \EndFor
 \end{algorithmic}
\end{algorithm}

\subsection{Convergence Analysis of ML-ISTA}

One can then inquire -- does the update in Equation \eqref{eq:MLISTA_update} provide convergent algorithm? Does the successive iterations minimize the original loss function? Though these questions have been extensively studied for proximal gradient methods \cite{beck2009fast,combettes2011proximal}, algorithms based on a proximal \emph{gradient mapping} have never been proposed -- let alone analyzed. Herein we intend to provide a first theoretical analysis of the resulting multi-layer thresholding approaches.

Let us formalize the problem assumptions, recalling that we are interested in
\begin{equation} \label{eq:MLlasso_general_2}
(P):\quad \min_{\gama}\ F(\gama) =   f(\D_2\gama) + g_1(\D_2\gama) + g_2(\gama),
\end{equation}
where $f:\mathbb{R}^{m_1} \to \mathbb{R}$ is a quadratic convex function, $g_1:\mathbb{R}^{m_1} \to \mathbb{R}$ is a convex and Lipschitz continuous function with constant $\ell_{g_1}$ and $g_2:\mathbb{R}^{m_2} \to (-\infty,\infty]$ is a proper closed and convex function that is $\ell_{g_2}$-Lipschitz continuous over its domain. Naturally, we will assume that both $g_1$ and $g_2$ are \emph{proximable}, in the sense that $\prox_{\alpha g_1}(\gama)$ and $\prox_{\alpha g_2}(\gama)$ can be efficiently computed for any $\gama$ and $\alpha>0$.
We will further require that $g_2$ has a bounded domain\footnote{This can be easily accommodated by adding to $g_2$ a norm bound constraint in the form of an indicator function $\delta_{B[0,R]}$, for some large enough $R>0$.}. More precisely, $\text{dom}(g_2) \subseteq \{\gama_2 : \|\gama_2 \|_2\leq R\}$. We denote $R_1 = \|\D_2\|_2R$, so that $\gama_2 \in \text{dom}(g_2) \implies \gama_1 = \D_2\gama_2$ satisfies $\|\gama_1\|_2 \leq R_1$.

Note also that the convex and smooth function $f$ can be expressed as $f(\gama_1) = \frac{1}{2}\gama_1^T\Q\gama_1 + \mathbf{b}^T\gama_1 + c$, for a positive semi-definite matrix $\Q = \D_1^T\D_1$. The gradient of $f$ can then easily be bounded by 
\begin{equation} \label{eq:M_bound}
    \|\nabla f(\gama_1)\|_2 \leq M \equiv  \|\Q\|_2 R_1 + \|\mathbf{b}\|_2,
\end{equation}
for any $\gama_1 = \D_2\gama_2$ with $\gama_2 \in \text{dom}(g_2)$.

It is easy to see that the algorithm in Equation \eqref{eq:ml_ista} does not converge to the minimizer of the problem $(P)$ by studying its fixed point. The point $\gama^\star_2$ is a fixed point of ML-ISTA if
\begin{align}
    & \exists  \ \w_2 \in\partial g_2(\gama^\star_2), \w_1 \in \partial g_1(\hat{\gama}_1) \\ & \text{such that} \ \D_2^T\nabla f(\D_2\gama^\star_2) + \D_2^T \w_1 +  \w_2 = 0,
\end{align}
where $\hat{\gama}_1 = \prox_{g_1\mu}(\D_2\gama^\star_2 - \mu \nabla f(\D_2 \gama^\star_2) )$. We extend on the derivation of this condition in Section \ref{app:fixed_point}. This is clearly different from the optimality conditions for problem $(P)$, which is 
\begin{align}
   & \exists \ \w_2 \in\partial g_2(\gama^\star_2), \w_1 \in \partial g_1(\D_2\gama_2^\star) \\ & \text{such that} \ \D_2^T\nabla f(\D_2\gama^\star_2) + \D_2^T \w_1 +  \w_2 = 0.
\end{align}
Nonetheless, we will show that the parameter $\mu$ controls the proximity of ``near''-fixed points to the optimal solution in the following sense: as $\mu$ gets smaller, the objective function of a fixed-point of the ML-ISTA method gets closer to $F_\text{opt}$, the minimal value of $F(\gama)$. In addition, for points that satisfy the fixed point equation up to some tolerance $\varepsilon>0$, the distance to optimality in terms of the objective function is controlled by both $\mu$ and $\varepsilon$. 

We first present the following lemma, stating that the norm of the gradient mapping operator is bounded. For clarity, we defer the proof of this lemma, as well as that of the following theorem, to Section \ref{Sec:proofs}.

\begin{lemma}\label{lemma:boundG} For any $\mu>0$ and $\gama_1 \in \mathbb{R}^{m_1}$, 
\begin{equation}
    \big\|G^{f,g_1}_{1/\mu}(\gama_1) \big\|_2\leq M + \ell_{g_1}.
\end{equation}
\end{lemma}

We now move to our main convergence result. In a nutshell, it states that the distance from optimality in terms of the objective function value of $\varepsilon$-fixed points is bounded by constants multiplying $\varepsilon$ and $\mu$.

\begin{theorem} \label{thm:main} 
Let\footnote{For a matrix $\A$, $\|\A\|_2$ denotes the spectral norm of $\A$: $\|\A\|_2 = \sqrt{\lambda_{\max}(\A^T\A)}$, where $\lambda_{\max}(\cdot)$ stands for the maximal eigenvalue of its argument.} $\mu \in \left(0,\frac{1}{\|\Q\|_2} \right)$, $t \in \left(0,\frac{4\mu}{3\|\D_2\|_2} \right)$, and assume $\gamat_2 \in \text{dom}(g_2)$ and $\gamat_1 = \D_2\gamat_2$. If
\begin{equation}
    \frac{1}{t} \left\| \gamat_2 - \prox_{tg_2} \left( \gamat_2 - t\ \D_2^T G^{f,g_1}_{1/\mu}(\gamat_1) \right) \right\|_2 \leq \varepsilon,
\end{equation}
then
\begin{equation}
    F(\alfa) - F_\text{opt} \leq \eta \varepsilon  + (\beta + \kappa t) \mu,
\end{equation}
where $F_\text{opt} \equiv \min_{\gama_2} F(\gama_2)$,
\begin{equation}
\alfa = \prox_{tg_2} \left( \gamat_2 - t \D_2^T G^{f,g_1}_{1/\mu}(\gamat_1) \right),    
\end{equation}
and 
\begin{align} \label{eq:Eta_Beta}
    \eta &= 2 R, \\
    \beta &= 2 R \|\D_2\|_2\|\Q\|_2(M+\ell_{g_1}) + \|\Q\|_2^2 R_1^2 \\ & \qquad \quad + 2\|\mathbf{b}\|_2 \|\Q\|_2 R_1 + \ell_{g_1}^2 + 2\ell_{g_1}M, \\
    \kappa &= \|\D_2\|_2 \left(\|\D_2\|_2 (M+\ell_{g_1}) + \ell_{g_2} \right) \|\Q\|_2 (M+\ell_{g_1}).
\end{align}
\end{theorem}

A consequence of this result is the following.

\begin{corollary}
Suppose $\{\gama_2^k\}$ is the sequence generated by ML-ISTA with $\mu\in\left( 0,\frac{1}{\|\D_1\|^2_2} \right)$ and $t\in\left( 0,\frac{4\mu}{3\|\D_2\|_2} \right)$. If $\|\gama_2^{k+1} - \gama_2^{k}\|_2\leq t\varepsilon$, then
\begin{equation}
    F(\gama_2^{k+1}) - F_\text{opt} \leq \eta \varepsilon  + (\beta + \kappa t) \mu,
\end{equation}
where $\eta$, $\beta$ and $\kappa$ are those given in \eqref{eq:Eta_Beta}.
\end{corollary}

An additional consequence of Theorem \ref{thm:main} is an analogous result for ML-FISTA. Recall that ML-FISTA introduces a momentum term to the update provided by ML-ISTA, and can be written as
\begin{align} \label{eq:ml-fista-update}
    \gama_2^{k+1} &= \prox_{tg_2} \left( \z^k - t \D_2^T G^{f,g_1}_{1/\mu}(\D_2\z^k) \right), \\
    \z^{k+1} &= \gama^{k+1}_2 + \rho^k (\gama_2^{k+1} - \gama_2^{k}).
\end{align}
We have the following result.

\begin{corollary} \label{crll:fista} 
Let $\mu \in \left(0,\frac{1}{\|\Q\|_2} \right)$ and $t \in \left(0,\frac{4\mu}{3\|\D_2\|_2} \right)$, and assume that $\{\gama_2^k\}$ and $\{\z^k\}$ are the sequences generated by ML-FISTA according to Equation \eqref{eq:ml-fista-update}. If
\begin{equation} \label{eq:assumption_thm2}
    \| \z^k - \gama_2^{k+1} \|_2 \leq t \varepsilon,
\end{equation}
then
\begin{equation}
    F(\gama_2^{k+1}) - F_\text{opt} \leq \eta \varepsilon  + (\beta + \kappa t) \mu,
\end{equation}
where the constants $\eta$, $\beta$ and $\kappa$ are defined in \eqref{eq:Eta_Beta}.
\end{corollary}

Before moving on, let us comment on the significance of these results. On the one hand, we are unaware of any results for proximal gradient-mapping algorithms, and in this sense, the analysis above presents a first result of its kind. On the other hand, the analysis does not provide a convergence rate, and so they do not reflect any benefits of ML-FISTA over ML-ISTA. As we will see shortly, the empirical convergence of these methods significantly differ in practice.

\begin{figure*}
\begin{center}
\includegraphics[width = .85\textwidth]{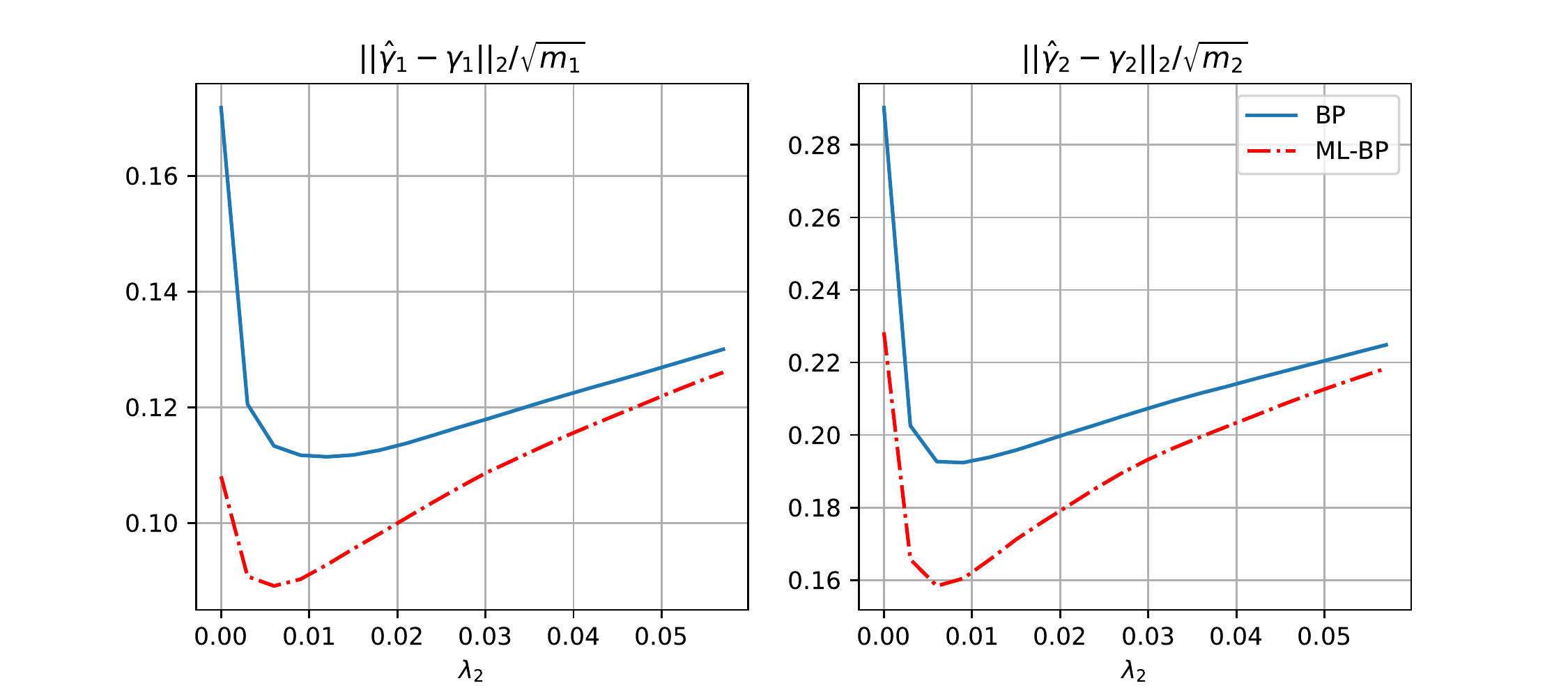}
\caption{Recovery error for $\gama_1$ and $\gama_2$ employing BP ($\lambda_1 = 0$) and Multi-Layer BP ($\lambda_1>0$).}
\label{fig:DemoML}
\end{center}
\end{figure*}

\begin{figure*}
\begin{center}
\includegraphics[width = .75\textwidth]{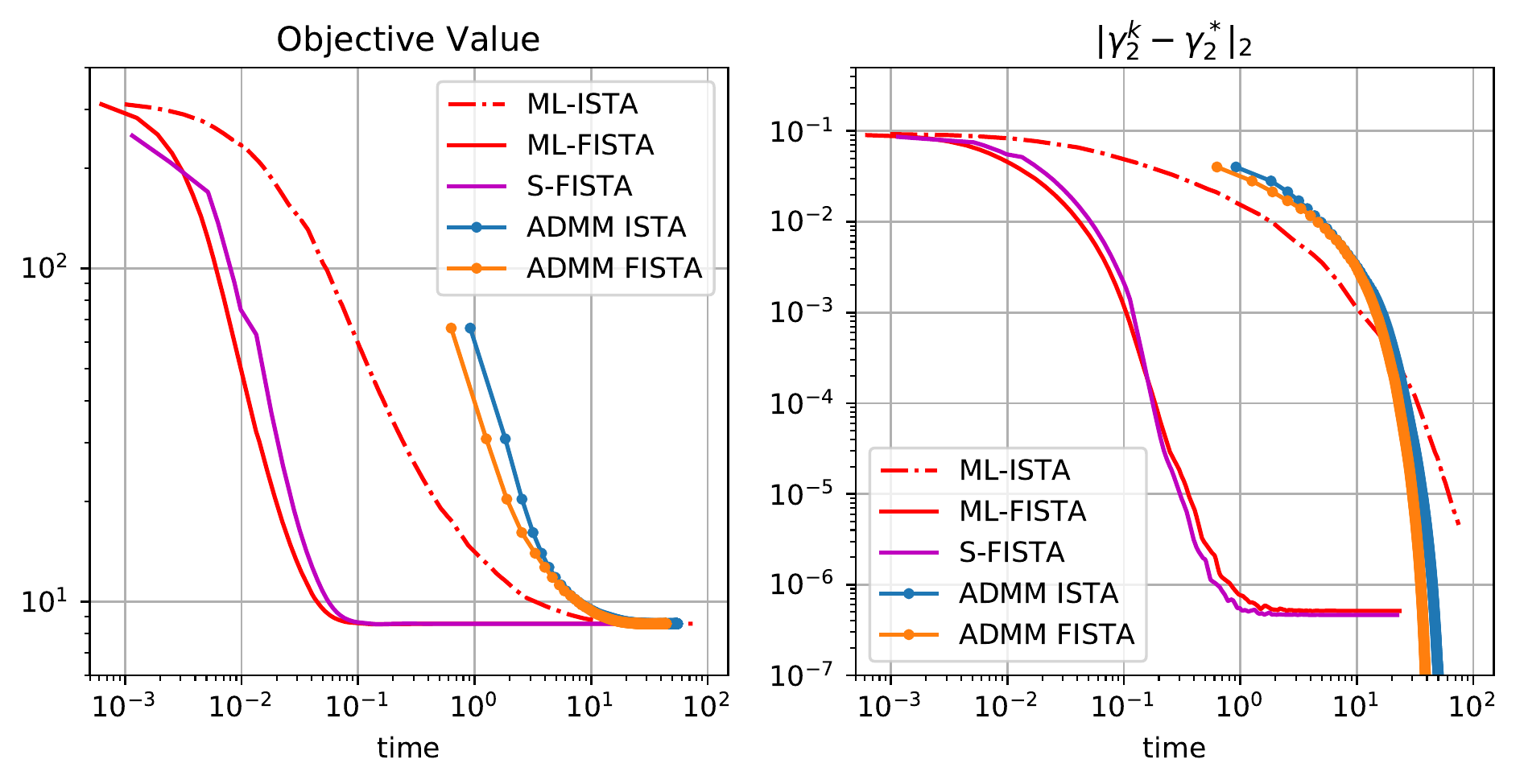}
\caption{Comparison of different solvers for Multi-Layer Basis Pursuit in terms of objective value (left) and distance to optimal solution (right).}
\label{fig:Synthetic_time}
\end{center}
\end{figure*}

\subsection{Synthetic Experiments}
We now carry a series of synthetic experiments to demonstrate the effectiveness of the multi-layer Basis Pursuit problem, as well as the proposed iterative shrinkage methods. 

\begin{figure*} \centering
\includegraphics[width = .75\textwidth]{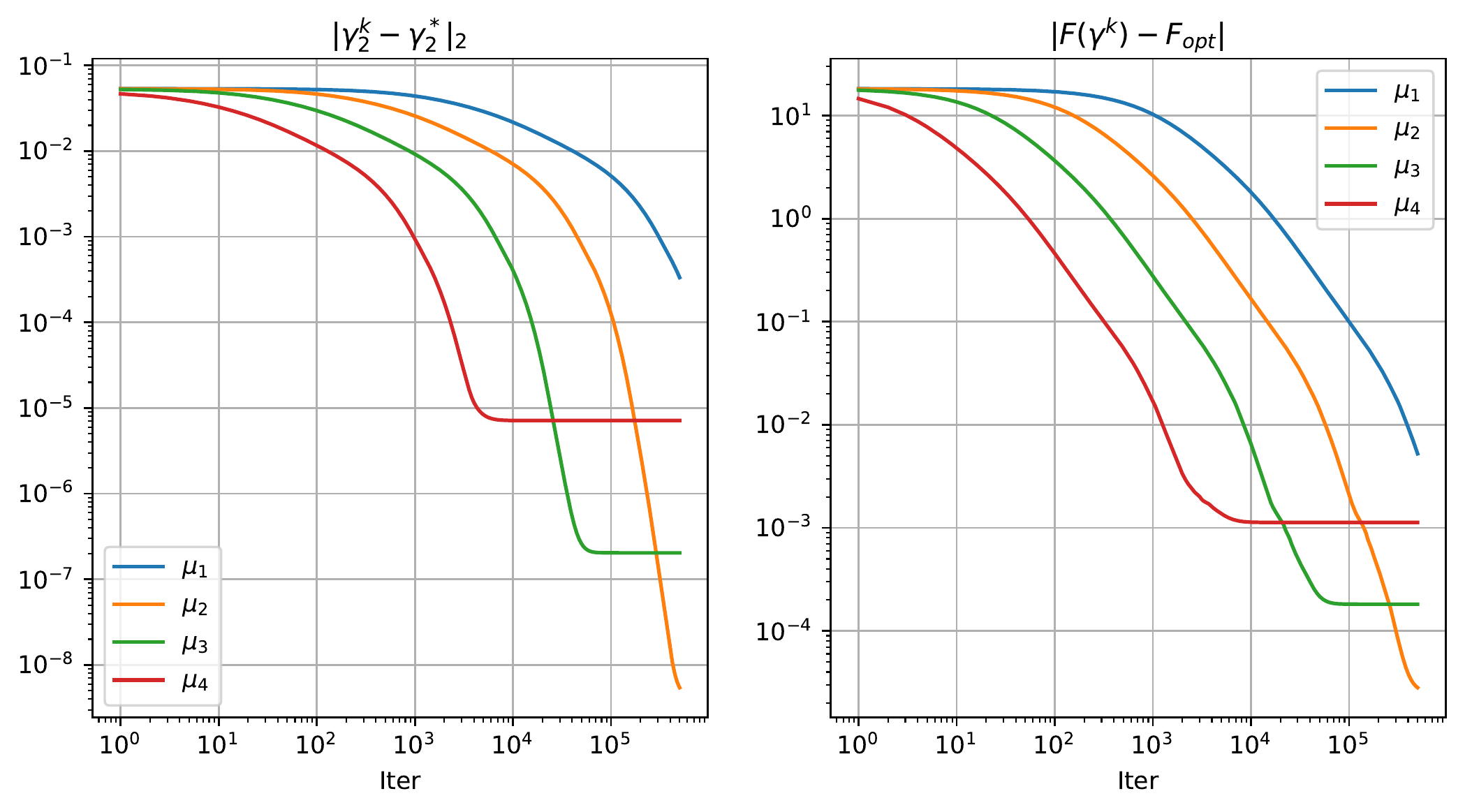} \\
\includegraphics[width = .75\textwidth]{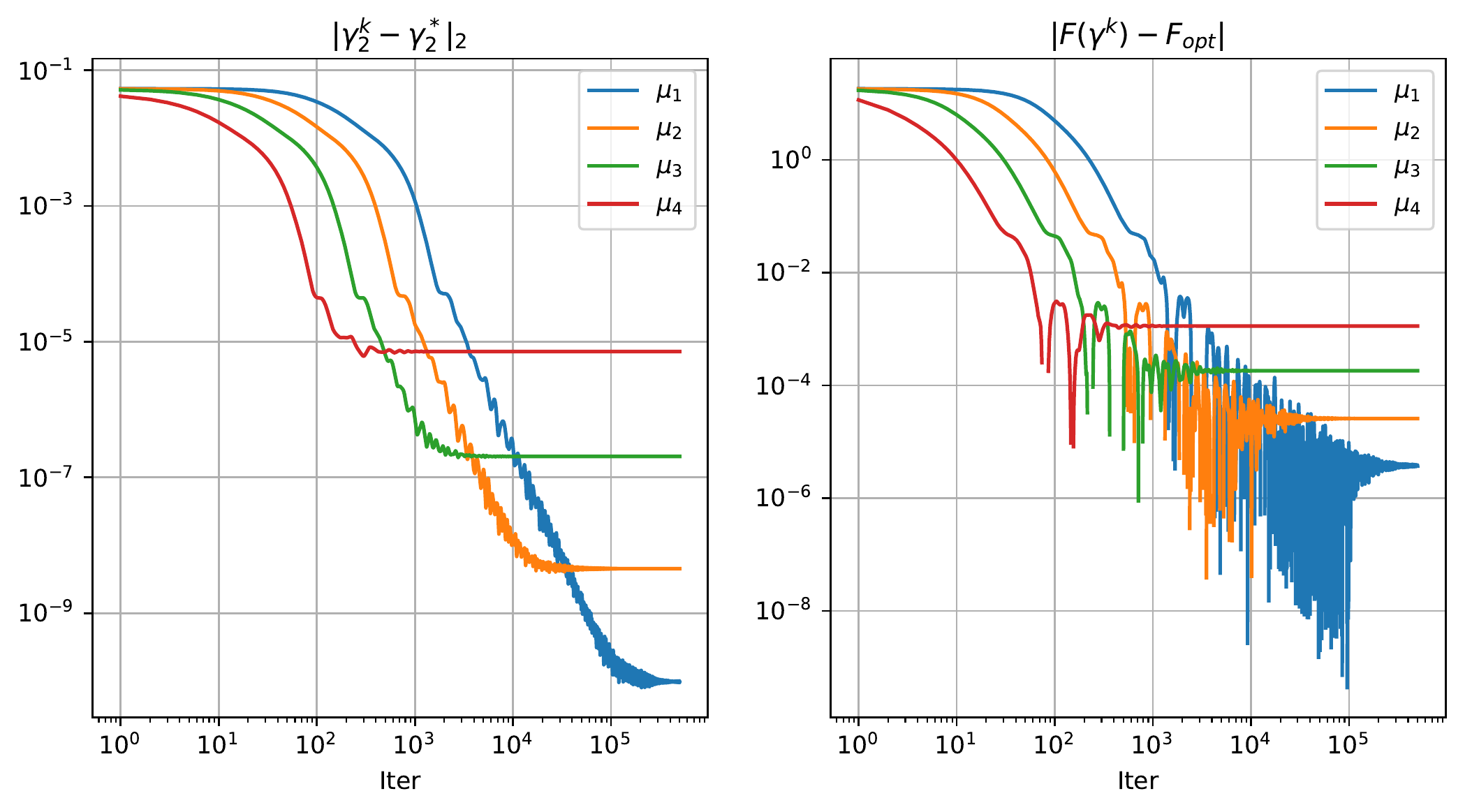}
\caption{ML-ISTA (top) and ML-FISTA (bottom) evaluation for different values of the parameter $\mu$.}
\label{fig:Synthetic_mu}
\end{figure*}
First, we would like to illustrate the benefit of the proposed multi-layer BP formulation when compared to the traditional sparse regression problem. In other words, exploring the benefit of having $\lambda_1 > 0$. To this end, we construct a two layer model with Gaussian matrices $\D_1 \in \mathbb{R}^{n\times m_1}$ and $\D_2 \in \mathbb{R}^{m_1\times m_2}$, ($n=50,m_1 = 70$, $m_2=60$). We construct our signals by obtaining representation $\gama_2$ with $\|\gama_2\|_0 = 30$ and $\|\gama_1\|_0 = 42$, following the procedure described in  \cite{Aberdam2018Holistic}, so as to provide representations consistent with the multi-layer sparse model. Lastly, we contaminate the signals with Gaussian i.i.d. noise creating the measurements $\y = \x + \w$ with SNR=10.
We compare minimizing $(P)$ with $\lambda_1=0$ (which accounts to solving a classical BP problem) with the case when $\lambda_1>0$, as a function of $\lambda_2$ when solved with ISTA and ML-ISTA. As can be seen from the results in Figure \ref{fig:DemoML}, enforcing the additional analysis penalty on the intermediate representation can indeed prove advantageous and provide a lower recovery error in both $\gama_2$ and $\gama_1$. 
Clearly, this is not true for any value of $\lambda_1$, as the larger this parameter becomes, the larger the bias will be in the resulting estimate. For the sake of this demonstration we have set $\lambda_1$ as the optimal value for each $\lambda_2$ (with grid search). The theoretical study of the conditions (in terms of the model parameters) under which $\lambda_1>0$ provides a better recovery, and how to determine this parameter in practice, are interesting questions that we defer to future work.

We also employ this synthetic setup to illustrate the convergence properties of the main algorithms presented above: ADMM (employing either ISTA or FISTA for the inner BP problem), the Smooth-FISTA \cite{beck2012smoothing} and the proposed Multi-Layer ISTA and Multi-Layer FISTA. Once again, we illustrate these algorithms for the optimal choice of $\lambda_1$ and $\lambda_2$ from the previous experiment, and present the results in Figure \ref{fig:Synthetic_time}. We measure the convergence in terms of function value for all methods, as well as the convergence to the solution found with ADMM run until convergence.

ADMM converges in relatively few iterations, though these take a relatively long time due to the inner BP solver. This time is reduced when using FISTA rather than ISTA (as the inner solver converges faster), but it is still significantly slower than any of the other alternatives - even while using warm-start at every iteration, which we employ in these experiments.

Both S-FISTA and our multi layer solvers depend on a parameter that controls the accuracy of their solution and affect their convergence speed --  the $\varepsilon$ for the former, and the $\mu$ for the latter approaches. We set these parameters so as to obtain roughly the same accuracy in terms of recovery error, and compare their convergence behavior. We can see that ML-FISTA is clearly faster than ML-ISTA, and slightly slightly faster than S-FISTA.
Lastly, in order to demonstrate the effect of $\mu$ in ML-ISTA and ML-FISTA, we run the same algorithms for different values of this parameter (in decreasing order and equispaced in logarithmic scale between $10^{-2.5}$ and 1) for the same setting, and present the results in Figure \ref{fig:Synthetic_mu} for ML-ISTA (top) and ML-FISTA (bottom). These numerical results illustrate the theoretical analysis provided by Theorem \ref{thm:main} in that the smaller $\mu$, the more accurate the solution becomes, albeit requiring more iterations to converge. These results also reflect the limitation of our current theoretical analysis, which is incapable of providing insights into the convergence rate.


\section{Principled Recurrent Neural Networks}

As seen above, the ML-ISTA and ML-FISTA schemes provide efficient solvers for problem $(P)$. Interestingly, if one considers the first iteration of either of the algorithms (with $\gama^0_L = \boldsymbol{0}$), the update of the inner most representation results in
\begin{equation} \label{eq:first_iter}
\gama_2 \gets  \frac{t}{\mu} \mathcal{T}_{t\lambda_2} \left( \D_2^T \mathcal{T}_{\mu\lambda_1}(\mu \D^T_1 ( \y))\right) ,
\end{equation}
for a two-layer model, for instance.
If one further imposes a non-negativity assumption on the representation coefficients, the thresholding operators $\mathcal{T}_\lambda$ become non-negative projections shifted by a bias of $\lambda$. Therefore, the above soft-thresholding operation can be equivalently written as 
\begin{equation}
 \gama_2 \gets \text{ReLU} \left( \D_2^T \text{ReLU}(\D_1^T \y + \mathbf{b}_1) + \mathbf{b}_2 \right)
\end{equation}
where the biases vectors $\mathbf{b}_1$ and $\mathbf{b}_2$ account for the corresponding thresholds\footnote{Note that this expression is more general in that it allows for different thresholds per atom, as opposed the expression in \eqref{eq:first_iter}. The latter can be recovered by setting every entry in the bias vector to be $\lambda_i$.}. Just as pointed out in \cite{papyan2016convolutional}, this is simply the forward pass in a neural network. Moreover, all the analysis presented above holds also in the case of convolutional dictionaries, where the dictionary atoms are nothing but convolutional filters (transposed) in a convolutional neural network. Could we benefit from this observation to improve on the performance of CNNs?

In this section, we intend to demonstrate how, by interpreting neural networks as approximation algorithms of the solution to a Multi-Layer BP problem, one can boost the performance of typical CNNs \emph{without introducing any parameters in the model}. To this end, we will first impose a generative model on the features $\gama_L$ in terms of multi-layer sparse convolutional representations; i.e., we assume that $\y \approx \D_{(1,L)}\gama_L$, for convolutional dictionaries $\D_i$.  Furthermore, we will adopt a supervised learning setting in which we attempt to minimize
an empirical risk over $N$ training samples of signals $\y_i$ with labels $h_i$. A classifier $\zeta_{\theta}(\gama^*)$, with parameters $\theta$, will be trained on estimates of said features $\gama^*(\y)$ obtained as the solution of the ML-BP problem; i.e.
\begin{multline} \label{eq:ClassProblem}
\min_{\theta,\{\D_i,\lambda_i\}} \frac{1}{N}\sum_{1=1}^{N} \mathcal{L}\left(h_i,\zeta_{\theta}(\gama^*)\right) \ \text{ s.t. } \\   \gama^* = \underset{\gama}{\arg\min}  \|\y - \D_{(1,L)}\gama\|^2_2 + \sum_{i=1}^{L-1}  \lambda_i\|\D_{(i+1,L)}\gama\|_1 + \lambda_L \|\gama\|_1.
\end{multline}
The function $\mathcal{L}$ is a loss or cost function to be minimized during training, such as the cross entropy which we employ for the classification case. Our approach to address this bi-level optimization problem is to approximate the solution of the lower-level problem by $k$ iterations of the ML-ISTA approaches -- effectively implemented as $k$ layers of unfolded recurrent neural networks. This way, $\gama^*$ becomes a straight-forward function of $\y$ and the model parameters ($\D_i$ and $\lambda_i$), which can be plugged into the loss function $\mathcal{L}$. A similar approach is employed by Task Driven Dictionary Learning \cite{mairal2012task} in which the constraint is a single layer BP (i.e. $L=1$) that is solved with LARS \cite{efron2004least} until convergence, resulting in a more involved algorithm.

\begin{figure*}
    \centering
    \includegraphics[width = .48\textwidth]{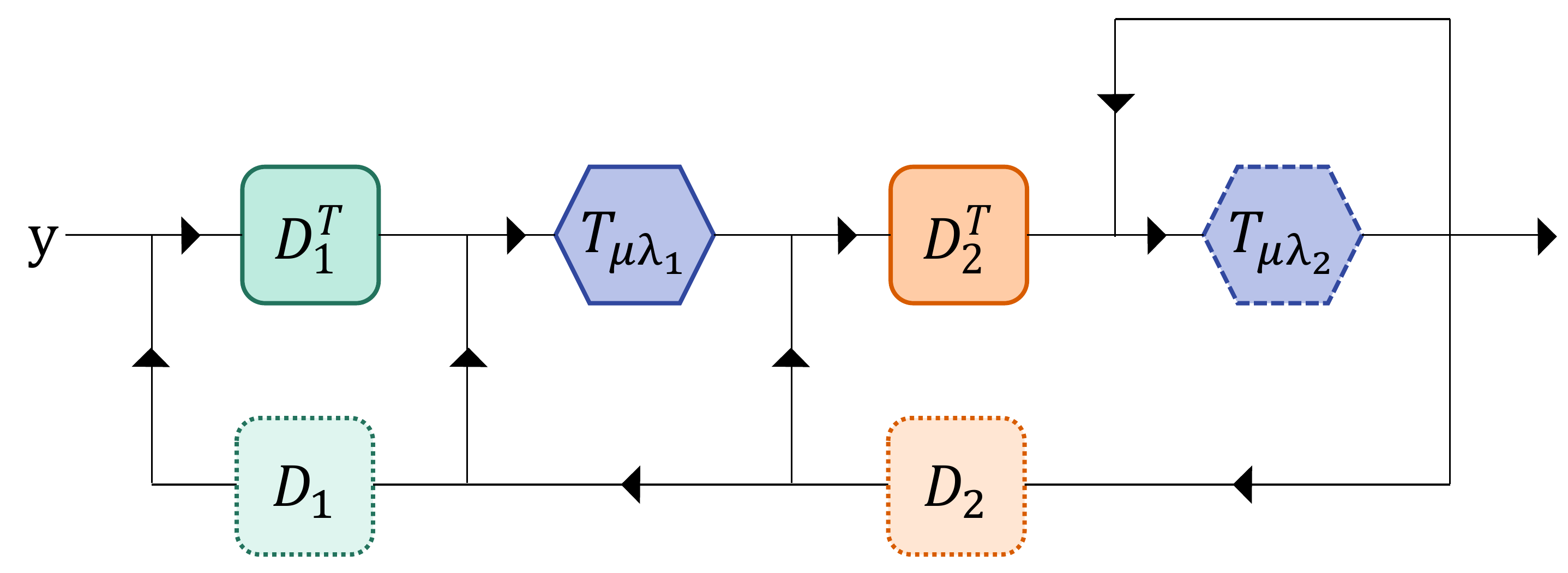} \\[.5cm] 
    \includegraphics[width = 1\textwidth]{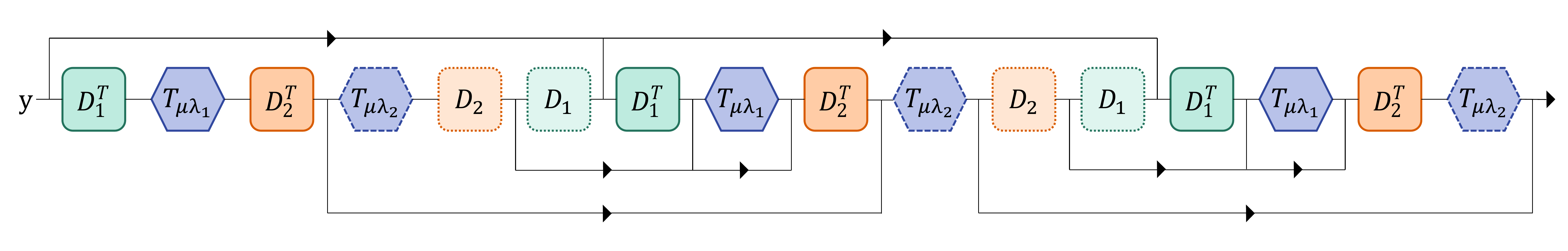} \\[.2cm]
    \caption{ML-ISTA graph interpretation for a two layer model as a recurrent neural network (top), and its unfolded version for 2 iterations (bottom).}
    \label{fig:ML-ISTA-graph}
\end{figure*}

Importantly, if only one iteration is employed for the ML-ISTA, and a linear classifier\footnote{Or, in fact, any other neural-network-based classifier acting on the obtained features.} is chosen for $\zeta_\theta(\gamma^*)$, the problem in \eqref{eq:ClassProblem} boils down exactly to training a CNN to minimize the classification loss $\mathcal{L}$. Naturally, when considering further iterations of the multi-layer pursuit, one is effectively implementing a recurrent neural network with ``skip connections'', as depicted in Figure \ref{fig:ML-ISTA-graph} for a two-layer model. These extended networks, which can become very deep, have exactly as many parameters as their traditional \emph{forward-pass} counterparts -- namely, the dictionaries $\D_i$, biases $\lambda_i$ and classifier parameters $\theta$. Notably, and unlike other popular constructions in the deep learning community (e.g., Residual Neural Networks \cite{he2016deep}, DenseNet \cite{huang2017densely}, and other similar constructions), these recurrent components and connections follow a precise optimization justification.

The concept of unfolding an iterative sparse coding algorithm is clearly not new. The first instance of such an idea was formalized by the Learned ISTA (LISTA) approach \cite{gregor2010learning}.
LISTA decomposes the linear operator of ISTA in terms of 2 matrices, replacing the computation of $\mathcal{T}_\lambda \left( \gama - \eta \D^T \left( \D \gama - \y \right) \right)$ by
\begin{equation}\label{eq:LISTA}
    \mathcal{T}_\lambda \left( \W \gama + \B \y \right), 
\end{equation}
following the equivalences $\W = \I - \eta \D^T\D$ and $\B = \D^T$. Then, it adaptively learns these new operators instead of the initial dictionary $\D$ in order to provide estimates $\hat{\gama}$ that approximate the solution of ISTA. Interestingly, such a decomposition allows for the acceleration of ISTA \cite{bruna2017UnderstandingTS}, providing an accurate estimate in very few iterations. A natural question is, then, could we propose an analogous multi-layer Learned ISTA? 

There are two main issues that need to be resolved if one is to propose a LISTA-like decomposition in the framework of our multi-layer pursuits. The first one is that the decomposition in \eqref{eq:LISTA} has been proposed and analyzed for general matrices (i.e., fully-connected layers in a CNN context), but not for convolutional dictionaries. If one was to naively propose to learn such an (unconstrained) operator $\W$, this would result in an enormous amount of added parameters. To resolve this point, in the case where $\D$ is a convolutional dictionary (as in CNNs) we propose a decomposition of the form
\begin{equation}\label{eq:ML-LISTA_1}
    \mathcal{T}_\lambda \left( (\I - \W^T\W) \gama + \B \y \right), 
\end{equation}
where $\W$ is also constrained to be convolutional, thus controlling the number of parameters\footnote{For completeness, we have also tested the \emph{traditional} decomposition proposed in Equation \eqref{eq:LISTA}, resulting in worse performance than that of ML-ISTA -- likely due to the significant increase in the number of parameters discussed above.}. In fact, the number of parameters in a layer of this ML-$L$ISTA is simply twice as many parameters as the conventional case, since the number of convolutional filters in $\W$ and $\B$ (and their dimensions) are equal to those in $\D$.

The second issue is concerned with the fact that LISTA was proposed as a relaxation of ISTA -- a pursuit tackling a \emph{single layer} pursuit problem. To accommodate a similar decomposition in our multi-layer setting, we naturally extend the update to:
\begin{align} \label{eq:update_mlLista}
    \hat{\gama_1} &\gets \mathcal{T}_{\lambda_1} \left( (\I - \W_1^T\W_1) \gama^k_1 + \B_1 \y \right), \\
    \gama^{k+1}_2 &\gets \mathcal{T}_{\lambda_2} \left( (\I - \W_2^T\W_2) \gama^k_2 + \B_2 \hat{\gama_1} \right),
\end{align}
for a two-layer model for simplicity. In the context of the supervised classification setting, the learning of the dictionaries $\D_i$ is replaced by learning the operators $\W_i$ and $\B_i$. Note that this decomposition prevents us from obtaining the dictionaries $\D_i$, and so we use\footnote{An alternative is to employ $\gama_1^k = \W_2 \gama^k_2$, but this choice was shown to perform slightly worse in practice.} $\gama_1^k = \B^T_2 \gama^k_2$ in Equation \eqref{eq:update_mlLista}.

\section{Experiments}

In this final section, we show how the presented algorithms can be used for image classification on three common datasets: MNIST, SVHN and CIFAR10, while improving the performance of CNNs \emph{without introducing any extra parameters in the model}. 
Recalling the learning formulation in Equation \eqref{eq:ClassProblem}, we will compare different architectures resulting from different solvers for the features $\gama^*$. As employing only one iteration of the proposed algorithms recovers a traditional feed-forward network, we will employ such a basic architecture as our baseline and compare it with the Multi Layer ISTA and FISTA, for different number of iterations or unfoldings. 
Also for this reason, we deliberately avoid using training ``tricks'' popular in the deep learning community, such as batch normalization, drop-out, etc., so as to provide clear experimental setups that facilitate the understanding and demonstration of the presented ideas.

For the MNIST case, we construct a standard (LeNet-style) CNN with 3 convolutional layers (i.e., dictionaries) with 32, 64 and 512 filters, respectively\footnote{Kernel sizes of $6\times 6$, $6\times 6$ and $4\times 4$, respectively, with stride of 2 in the first two layers.}, and a final fully-connected layer as the classifier $\zeta(\gama^*)$. We also enforce non-negativity constraints on the representations, resulting in the application of ReLUs and biases as shrinkage operators. For SVHN we use an analogous model, though with three input channels and slightly larger filters to accommodate the larger input size. For CIFAR, we define a ML-CSC model with 3 convolutional layers, and the classifier function $\zeta(\gama^*)$ as a three-layer CNN. This effectively results in a 6 layers architecture, out of which the first three are unfolded in the context of the multi-layer pursuits. All models are trained with SGD with momentum, decreasing the learning rate every so many iterations. In particular, we make use of a PyTorch implementation, and training code is made available\footnote{Available through the first author's website.} online.

\begin{figure} \centering
\includegraphics[width = .4\textwidth]{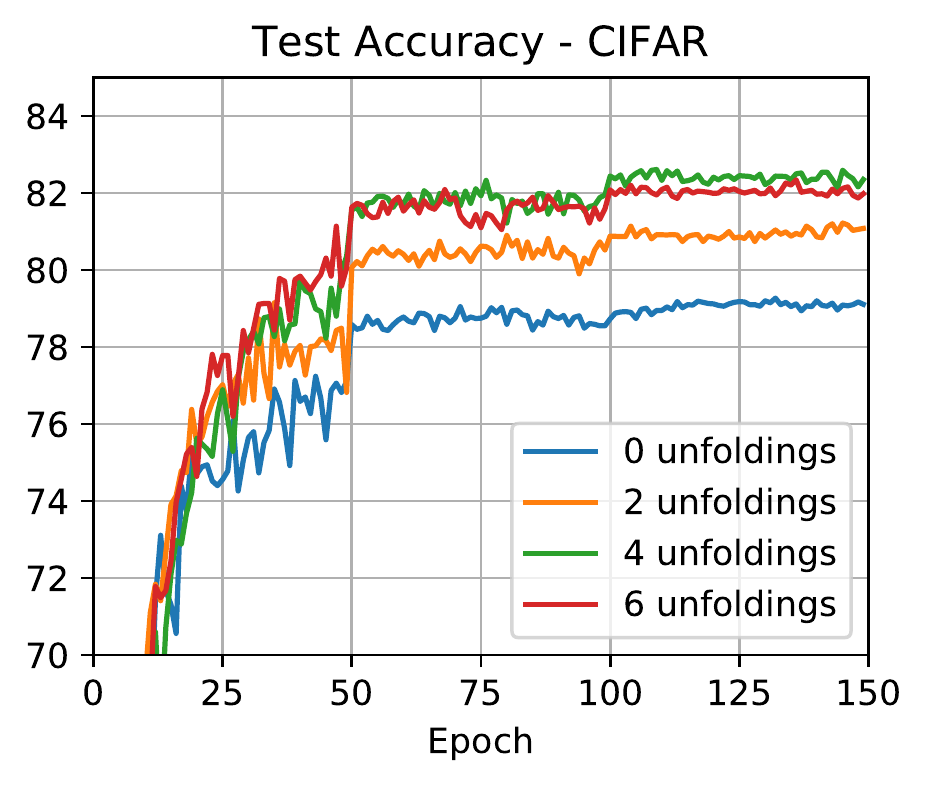}
\caption{Training ML-ISTA for different number of unfoldings, on CIFAR10. The case of 0 unfoldins corresponds to the traditional feed-forward convolutional network. All networks have the same number of parameters.}
\label{fig:unfoldings}
\end{figure}

In order to demonstrate the effect of the ML-ISTA iterations (or unfoldings), we first depict the test error as a function of the training epochs for different number of such iterations in Figure \ref{fig:unfoldings}. Recall that the case of 0 unfoldings corresponds to the typical feed-forward CNN, while the case with 6 unfoldings effectively implements a 18-layers-deep architecture, alas having the same number of parameters. As can be seen, further unfoldings improve on the resulting performance. 

Moving to a more complete comparison, we demonstrate the ML-ISTA and ML-FISTA architectures when compared to some of the models mentioned above; namely:
\begin{itemize}
    \item \emph{ML-LISTA}: replacing the learning of the convolutional dictionaries (or filters) by the learning of the (convolutional) factors $\W_i$ and $\B_i$, as indicated in Equation \eqref{eq:update_mlLista}.
    \item \emph{Layered Basis Pursuit}: the approach proposed in \cite{papyan2016convolutional}, which unrolls the iteration of ISTA for a single-layer BP problem \emph{at each layer}. In contrast, the proposed ML-ISTA/FISTA unrolls the iterations of the entire Multi-Layer BP problem. 
    \item An ``\emph{All-Free}'' model: What if one ignores the generative model (and the corresponding pursuit interpretation) and simply frees all the filters to be adaptively learned? In order to study this question, we train a model with the same depth and an analogous recurrent architecture as the unfolded ML-ISTA/FISTA networks, but where all the filters of the different layers are free to be learned and to provide the best possible performance.
\end{itemize}
It is worth stressing that the ML-ISTA, ML-FISTA and Layered BP have all the same number of parameters as the feed-forward CNN. The ML-LISTA version has twice as many parameters, while the All-Free version has order $\mathcal{O}(LK)$  more parameters, where $L$ is the number of layers and $K$ is the number of unfoldings.

\begin{figure*}
\includegraphics[ width=.335\textwidth]{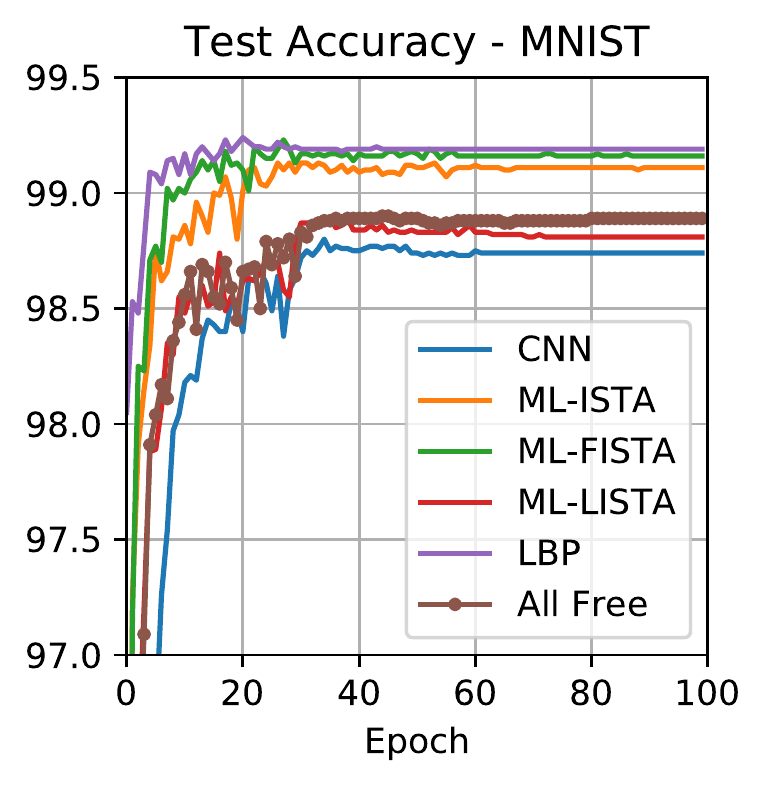}
\includegraphics[ width=.32\textwidth]{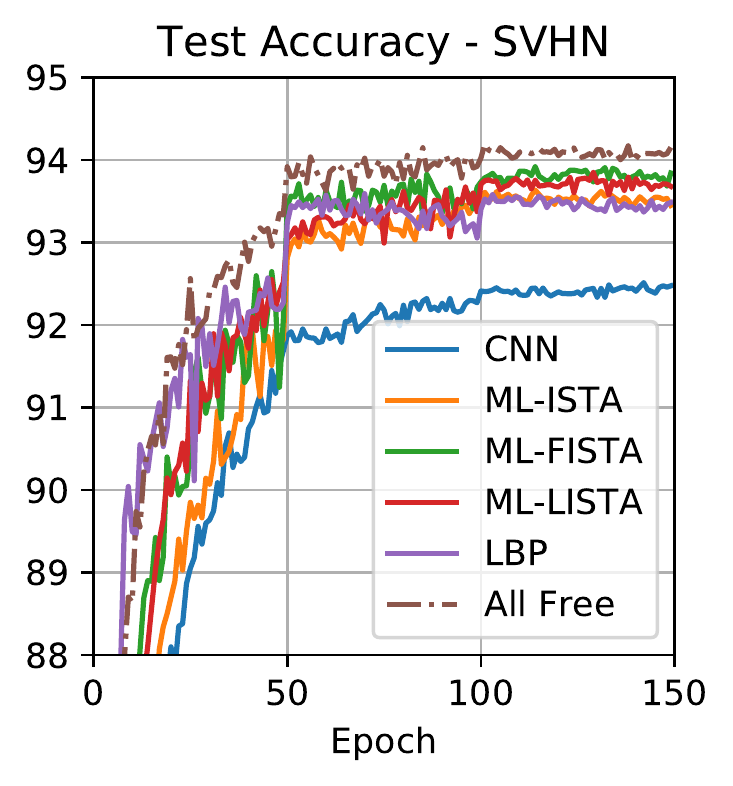}
\includegraphics[ width=.32\textwidth]{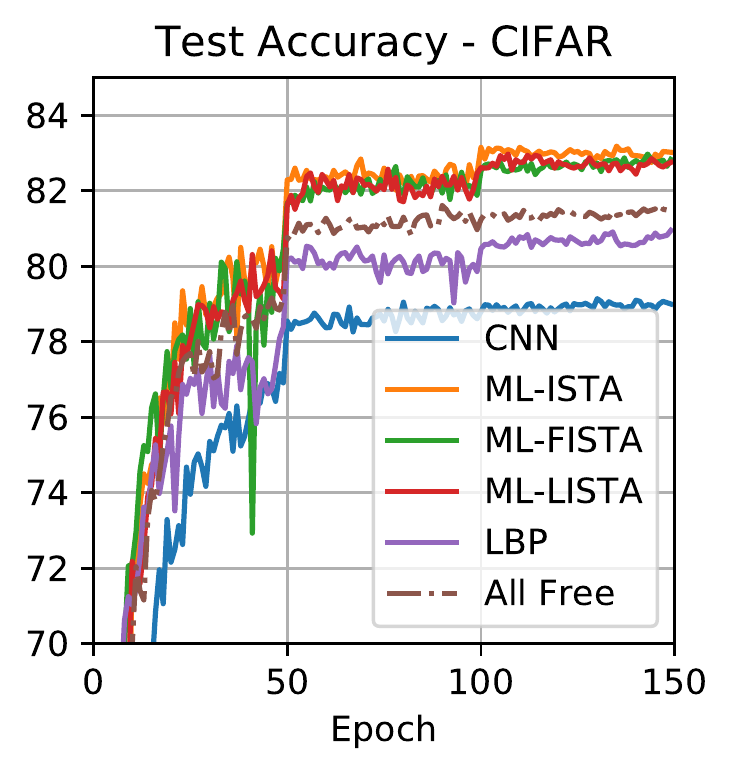}
\caption{Comparison of different architectures on the SVHN dataset, with a feed-forward network as baseline. All networks have the same number of parameters.}
\label{fig:ClassificationResults}
\end{figure*}

The accuracy as a function of the iterations for all models are presented in Figure \ref{fig:ClassificationResults}, and the final results are detailed in Table \ref{table:class_resuts}. A first observation is that most ``unrolled'' networks provide an improvement over the baseline feed-forward architecture. Second, while the Layered BP performs very well on MNIST, it falls behind on the other two more challenging datasets. Recall that while this approach unfolds the iterations of a pursuit, it does so one layer at a time, and does not address a global pursuit problem as the one we explore in this work. 

Third, the performances of the ML-ISTA, ML-FISTA and ML-LISTA are comparable. This is interesting, as the LISTA-type decomposition does not seem to provide an importance advantage over the unrolled multi-layer pursuits. 
Forth, and most important of all, freeing all the parameters in the architecture does not provide important improvements over the ML-ISTA networks. Limited training data is not likely to be the cause, as ML-ISTA/FISTA outperforms the larger model even for CIFAR, which enjoys a rich variability in the data and while using data-augmentation. This is noteworthy, and this result seems to indicate that the consideration of the multi-layer sparse model, and the resulting pursuit, does indeed provide an (approximate) solution to the problem behind CNNs.

\begin{table}[h]
		\centering
		\begin{tabular}{ l c c c }
			\toprule
			Model & MNIST & SVHN & CIFAR 10 \\
			\midrule
			Feed-Forward               & 98.78 \%  & 92.44 \% &  79.00 \% \\
			Layered BP 				   & 99.19 \%  & 93.42 \%  & 80.73 \% \\ 
			ML-ISTA                    & 99.10 \%  & 93.52 \%  & 82.93 \% \\
			ML-FISTA                   & 99.16 \%  & 93.79 \% &  82.79 \% \\  
			ML-LISTA                   & 98.81 \%  & 93.71 \% &  82.68 \% \\  
			All-Free                   & 98.89 \%  & 94.06 \% &  81.48 \% \\  
			\bottomrule \vspace{.1cm}
		\end{tabular}
		\caption{Classification results for different architectures for MNIST, SVHN and CIFAR10.}
		\label{table:class_resuts}
\end{table}

\section{Proofs of Main Theorems}
\label{Sec:proofs}
\subsection{Fixed Point Analysis}

\label{app:fixed_point}
A vector $\gama_2^\star$ is a fixed point of the ML-ISTA update from Equation \eqref{eq:ml_ista} iff
\begin{equation}
\gama_2^\star = \prox_{tg_2} \left( \gama_2^\star - t\ \D_2^T\  G_{1/\mu}^{f,g_1}(\D_2\gama_2^\star) \right).
\end{equation}
By the second prox theorem \cite[Theorem 6.39]{B17}, we have that
\begin{equation}
    - t\ \D_2^T\  G_{1/\mu}^{f,g_1}(\D_2\gama_2^\star) \in t \partial g_2(\gama_2^\star),
\end{equation}
or, equivalently, there exists $\w_2 \in \partial g_2(\gama_2^\star)$ so that
\begin{equation}
     \D_2^T\  G_{1/\mu}^{f,g_1}(\D_2\gama_2^\star) + \w_2 = 0.
\end{equation}
Employing the definition of $G_{1/\mu}^{f,g_1}(\D_2\gama_2^\star)$,
\begin{equation}\label{eq:first}
     \D_2^T\ \frac{1}{\mu} \left( \D_2 \gama_2^\star - \prox_{\mu g_1}(\D_2 \gama_2^\star - \mu \nabla f(\D_2 \gama_2^\star) ) \right) + \w_2 = 0.
\end{equation}
Next, denote 
\begin{equation}\label{eq:app_gamma_1_hat}
\hat{\gama}_1 = \prox_{\mu g_1}(\D_2 \gama_2^\star - \mu \nabla f(\D_2 \gama_2^\star) ).
\end{equation} 
Employing the second prox theorem on \eqref{eq:app_gamma_1_hat}, we have that the above is equivalent to the existence of $\w_1 \in \partial g_1 (\hat{\gama}_1)$ for which $\D_2\gama_2^\star - \mu \nabla f(\D_2\gama_2^\star)-\hat{\gama}_1 = \mu \w_1$. Thus, \eqref{eq:first} amounts to 
\begin{equation}
     \D_2^T\ \frac{1}{\mu} \left( \D_2 \gama_2^\star -      \D_2\gama_2^\star + \mu \nabla(\D_2\gama_2^\star) + \mu \w_1        \right) + \w_2 = 0,
\end{equation}
for some $\w_2 \in \partial g_2 (\gama^\star_2)$ and  $\w_1 \in \partial g_1 (\hat{\gama}_1)$. Simplifying the expression above we arrive at the fixed-point condition of ML-ISTA, which is 
\begin{align}
    & \exists \ \w_2 \in\partial g_2(\gama^\star_2), \w_1 \in \partial g_1(\hat{\gama}_1) \\ &\text{so that} \ \D_2^T\nabla f(\D_2\gama^\star_2) + \D_2^T \w_1 +  \w_2 = 0.
\end{align}

\subsection{Proof of Lemma \ref{lemma:boundG}} 

\label{app:proof_Lemma}

\begin{proof}
Denote $S_\mu(\gama_1) = \prox_{\mu g_1}(\gama_1 - \mu \nabla f(\gama_1))$. Then, by the second prox theorem (\cite[Theorem 6.39]{B17}), we have that
\begin{equation}
    \gama_1 - \mu \nabla f(\gama_1) - S_\mu(\gama_1) \ \in \ \mu \partial g_1(S_\mu(\gama_1)).
\end{equation}
Dividing by $\mu$ and employing the definition of the gradient mapping, we obtain
\begin{equation}
    G_{1/\mu}^{f,g_1}(\gama_1) \in \nabla f(\gama_1) + \partial g_1(S_\mu(\gama_1)).
\end{equation}
By the $\ell_{g_1}$-Lipschitz continuity of $g_1$ \cite[Theorem 3.61]{B17}, it follows that $\|\z\|_2 \leq \ell_{g_1}$ for any $\z \in \partial g_1(S_\mu(\gama_1))$. This, combined with the bound $\|\nabla f(\gama_1) \|_2 \leq M$ from Equation \eqref{eq:M_bound}, provides the desired claim.
\end{proof}

\subsection{Proof of Theorem \ref{thm:main}} \label{app:proof_main}

\def\Gfgmu{G^{f,g_1}_{1/\mu}}
\def\a{\mathbf{a}}
\def\gamast{\gama_\mu^\ast}

\begin{proof}
Denote
\begin{align} \label{eq:definition_of_a's}
    \a_1 &= \frac{1}{t} \left[ \gamat_2 - \prox_{tg_2} \left( \gamat_2 - t\D_2^T(\I - \mu \Q) \Gfgmu(\gamat_1) \right) \right], \\
    \a_2 &= \frac{1}{t} \left[ \gamat_2 - \prox_{tg_2} \left( \gamat_2 - t \D_2^T \Gfgmu(\gamat_1) \right) \right].
\end{align}
By the triangle inequality,
\begin{equation} \label{eq:IneqA1}
    \|\a_1\|_2 \leq \|\a_2\|_2 + \|\a_1-\a_2\|_2.
\end{equation}
We will upper-bound the right-hand side of this inequality. First, employing the non-expansiveness property of prox operators (\cite[Lemma 2.4]{CW05}), we can write
\begin{align}
    \| \a_1 - \a_2 \|_2 = &\ \frac{1}{t} \Big\| \prox_{tg_2} \left( \gamat_2 - t\D_2^T \Gfgmu (\gamat_1) \right) \\ & - \prox_{tg_2} \left( \gamat_2 - t\D_2^T (\I - \mu \Q) \Gfgmu (\gamat_1) \right) \Big\|_2 \\
    \leq &\ \Big\| \D_2^T (\I - \mu \Q) \Gfgmu (\gamat_1) - \D_2^T \Gfgmu (\gamat_1) \Big \|_2 \\
    = &\ \mu \Big\| \D_2^T \Q \ \Gfgmu(\gamat_1) \Big\|_2 \\ \label{eq:a1-a2}
    \leq &\ \mu \|\D_2\|_2 \|\Q\|_2 (M+\ell_{g_1}),
\end{align}
where the last inequality follows from the definition of operator norms and Lemma \ref{lemma:boundG}. Also, $\| \a_2 \|_2 \leq \varepsilon$ by assumption. 
Thus, from \eqref{eq:IneqA1},
\begin{equation} \label{eq:bound_a_1}
    \|\a_1\|_2 \leq \varepsilon + \mu \|\D_2\|_2 \|\Q\|_2 (M+\ell_{g_1}).
\end{equation}

Consider now the function $H_\mu: \mathbb{R}^{m_1} \to \mathbb{R}$ given by
\begin{multline}
    H_\mu(\gama) =  \frac{1}{2}\gama^T (\Q - \mu \Q^2) \gama + \mathbf{b}^T (\I - \mu \Q)\gama \\ + M^\mu_{g_1}((\I-\mu \Q)\gama - \mu \mathbf{b}), 
\end{multline}
where $M^\mu_{g_1}$ is the Moreau envelope of $g_1$ with smoothness parameter $\mu$ \cite{M65}. Note that $H_\mu$ is convex since $\mu<\frac{1}{\|\Q\|_2}$ implies $\Q - \mu \Q^2 \succeq 0$ and the Moreau envelope of a convex function is convex. Recall that the gradient of the Moreau envelop is given by $\nabla M^\mu_{g_1}(\gama) = \frac{1}{\mu}(\gama - \prox_{\mu g_1}(\gama))$ (see e.g. \cite[Theorem 6.60]{B17}), and so
\begin{align}
    \nabla H_\mu(\gama) =& \ \frac{1}{\mu} (\I - \mu\Q)\left[ \gama - \prox_{\mu g_1} (\gama-\mu(\Q\gama+\mathbf{b})) \right] \\
    =& \ (\I - \mu \Q)\Gfgmu(\gama).
\end{align}
Consider now $\gama_1 = \D_2\gama_2$ and $\tilde{H}_\mu(\gama_2) \equiv H_\mu(\D_2\gama_2)$. Applying the chain rule yields 
\begin{align}
   \nabla \tilde{H}_\mu(\gama_2)  =  \D_2^T(\I - \mu \Q)\Gfgmu(\gama_1).
\end{align}
Thus, we can conclude that $\a_1$ is nothing else than the gradient mapping of $\tilde{H}_\mu$ and $g_2$, and so the inequality in \eqref{eq:bound_a_1} can be rewritten as
\begin{equation}\label{eq:bound_G_H}
    \left\|G^{\tilde{H}_\mu,g_2}_{1/t}(\gamat_2)\right\|_2 \leq \varepsilon + \mu \|\D_2\|_2 \|\Q\|_2 (M+\ell_{g_1}).
\end{equation}
Gradient mapping operators are firmly non-expansive with constant $\frac{3\mu}{4}$ (\cite[Lemma 10.11]{B17}), from which it follows that $H_\mu$ is $\frac{4}{3\mu}$-smooth. Denote $F_\mu(\gama_2) = H_\mu(\D_2\gama_2) + g_2(\gama_2)$, and one of its minimizers by $\gama_\mu^\ast \in \arg\min F_\mu(\gama_2)$. Moreover, define 
\begin{equation}
    \hat{\gama} = \prox_{tg_2} \left( \gamat_2 - t \D_2^T (\I-\mu\Q) G^{f,g_1}_{1/\mu}(\gamat_1) \right).
\end{equation}
By the fundamental prox-grad inequality (\cite[Theorem 10.16]{B17}), and since $t \in \left( 0 , \frac{4\mu}{3\|\D_2\|_2} \right)$, it follows that
\begin{equation}
    F_\mu(\gamast)-F_\mu(\hat{\gama}) \geq \frac{1}{2t} \|\gamast - \hat{\gama}\|_2^2 - \frac{1}{2t}  \| \gamast - \gamat_2 \|^2_2.
\end{equation}
Then, by the three-points lemma (see \cite{CT93}), we may rewrite
\begin{equation}
    \|\gamast - \hat{\gama}\|_2^2 - \| \gamast - \gamat_2 \|^2_2 = 2 \langle \gamast - \hat{\gama}, \gamat_2 - \hat{\gama} \rangle - \| \gamat_2 - \hat{\gama} \|^2_2.
\end{equation}
Thus,
\begin{align}
    F_\mu(\hat{\gama})-F_\mu(\gamast) &\leq \frac{1}{t} \langle \hat{\gama} - \gamast ,\gamat_2 - \hat{\gama} \rangle + \frac{1}{2t}\| \gamat_2 - \hat{\gama} \|^2_2 \\
    &= \frac{1}{t} \langle \hat{\gama} - \gamat_2 , \gamat_2 - \hat{\gama} \rangle \\ & \quad + \frac{1}{t} \langle \gamat_2 - \gamast , \gamat_2 - \hat{\gama} \rangle  + \frac{1}{2t}\| \gamat_2 - \hat{\gama} \|^2_2 \\
    &= -\frac{1}{2t} \| \gamat_2 - \hat{\gama} \|_2^2 + \frac{1}{t} \langle \gamat_2 - \gamast , \gamat_2 - \hat{\gama} \rangle \\
    &\leq \langle \gamat_2 - \gamast , G_{1/t}^{\tilde{H}_\mu,g_2}(\gamat_2) \rangle \\
    &\leq 2 R \|G_{1/t}^{\tilde{H}_\mu,g_2}(\gamat_2)\|_2,
\end{align}
where the last passage uses the Cauchy-Schwarz inequality along with $\|\gamat_2 - \gamast\|_2\leq 2R$. Combining the above with Inequality \eqref{eq:bound_G_H} yields
\begin{equation} \label{eq:Dif_Fs}
    F_\mu(\hat{\gama})-F_\mu(\gamast) \leq 2R\varepsilon + 2\mu R \|\Q\|_2\|\D_2\|_2(M + \ell_{g_1}).
\end{equation}
Finally, we will connect between $F_\mu(\gamast)$, $F_\mu(\gamat_2)$ and $F(\gamast)$, $F(\gamat_2)$, respectively. Note that for any $\gama_2\in\text{dom}(g_2)$ and $\gama_1 = \D_2\gama_2$,
\begin{multline}
    \Big| H_\mu(\gama_1) - f(\gama_1) - g_1(\gama_1) \Big| =
    \Big| -\mu \left( \frac{1}{2} \|\Q\gama_1 \|_2^2 + \mathbf{b}^T\Q\gama_1 \right) \\ + M^\mu_{g_1} \left( (\I - \mu \Q)\gama_1 - \mu\mathbf{b} \right) - g_1(\gama_1) \Big|.
\end{multline}
Moreover, this expression can be upper-bounded by
\begin{multline}
     \left( \frac{1}{2}\|\Q\|^2_2 R_1^2 + \|\mathbf{b}\|_2\|\Q\|_2 R_1 \right) \mu  \\ + \left| M^\mu_{g_1}\left( (\I - \mu \Q)\gama_1 - \mu\mathbf{b} \right) - g_1(\gama_1) \right|.
\end{multline}
Further, from basic properties of the Moreau envelope (in particular Theorem 10.51 in \cite{B17}) and the $\ell_{g_1}$-Lipschitz property of $g_1$, we have that
\begin{multline}
     \big| M^\mu_{g_1} \left( (\I - \mu \Q)\gama_1 - \mu\b \right) - g_1(\gama) \big| \leq \\ \qquad 
     \big| M^\mu_{g_1}\left( (\I - \mu \Q)\gama_1 - \mu\b \right) - g_1((\I - \mu\Q)\gama_1-\b) \big| \\  \qquad \qquad + \big| g_1((\I - \mu\Q)\gama_1-\b) - g_1(\gama_1) \big| \\
     \leq  \frac{\ell_{g_1}^2}{2} \mu + \ell_{g_1}M\mu.
\end{multline}
We thus obtain that for any $\gama_2\in\text{dom}(g_2)$
\begin{equation} \label{eq:Fmu-F}
    \Big| H_\mu(\gama_1) - f(\gama_1) - g_1(\gama_1) \Big| = \big| F_\mu(\gama_2) - F(\gama_2) \big| \leq C\mu,
\end{equation}
where 
\begin{equation}
    C = \frac{R_1^2}{2}\|\Q\|^2_2 + \|\mathbf{b}\|_2\|\Q\|_2R_1 + \frac{\ell_{g_1}^2}{2} + \ell_{g_1}M.
\end{equation}
From this, we have that
\begin{align}\label{eq:ineq_2}
    F(\hat{\gama})\leq & F_\mu(\hat{\gama}) + C\mu.
\end{align}
Recall now that $\alfa = \prox_{tg_2} \left( \gamat_2 - t \D_2^T G^{f,g_1}_{1/\mu}(\gamat_1) \right)$. Then,
\begin{align}
    | F(\alfa) - F(\hat{\gama}) | \leq & | f(\D_2\alfa) - f(\D_2\hat{\gama}) | \\ & + | g_1(\D_2\alfa) - g_1(\D_2\hat{\gama})| \\ & + | g_2(\alfa) - g_2(\hat{\gama}) | \\ \label{eq:ContinuityF}
    \leq & \left( \|\D_2\|_2 ( M + \ell_{g_1}) + \ell_{g_2} \right) \| \alfa - \hat{\gama} \|_2,
\end{align}
where we have used the Lipschitz continuity of $g_1$ and $g_2$. Next, note that $\| \alfa - \hat{\gama} \|_2 = t \|\a_1 - \a_2\|$ from \eqref{eq:a1-a2}. This way,
\begin{equation} \label{eq:Falfa<Fgama}
    F(\alfa) \leq F(\hat{\gama}) + t \kappa \mu,
\end{equation}
where $\kappa = \|\D_2\|_2 \left(\|\D_2\|_2 ( M + \ell_{g_1}) +   \ell_{g_2} \right) \|\Q\|_2 (M+\ell_{g_1})$.

Returning to \eqref{eq:Fmu-F}, and because 
\begin{equation}
    \min_{\gama_2} F(\gama_2) \geq \min_{\gama_2} F_\mu(\gama_2) - C\mu,
\end{equation}
we have that
\begin{equation}\label{eq:ineq_3}
    F_\text{opt} \geq F_\mu(\gamast) - C\mu.
\end{equation}
Finally, combining \eqref{eq:Dif_Fs}, \eqref{eq:Falfa<Fgama} and \eqref{eq:ineq_3}, we obtain
\begin{equation}
    F(\alfa) - F_\text{opt} \leq 2R\varepsilon + t \mu \kappa +  (2R\|\D_2\|_2\|\Q\|_2(M+\ell_{g_1}) + 2C ) \mu.
\end{equation}
\end{proof}

\section{Conclusion}
Motivated by the multi-layer sparse model, we have introduced a multi-layer basis pursuit formulation which enforces an $\ell_1$ penalty on the intermediate representations of the ML-CSC model. We showed how to solve this problem effectively through  multi-layer extensions of iterative thresholding algorithms, building up on a projected gradient mapping approach. We showed that $\varepsilon$-fixed points provide approximations that are arbitrarily close, in function value, to the optimal solution. Other theoretical questions, such as those of convergence rates, constitute part of ongoing research.

We further showed how these algorithms generalize feed-forward CNN architectures by principled residual ones, improving on their performance as subsequent iterations are considered. It is intriguing how one could employ the recent results in \cite{giryes2018tradeoffs,bruna2017UnderstandingTS} to the analysis of our resulting unfolded networks, or to understand why the Learned ML-LISTA does not provide further benefits over ML-ISTA/FISTA in the studied cases. More broadly, we believe that the study and analysis of these problems will likely contribute to the further understanding of deep learning.

\bibliographystyle{IEEEtran}
\bibliography{mybib}

\begin{thebibliography}{10}
\providecommand{\url}[1]{{#1}}
\providecommand{\urlprefix}{URL }
\expandafter\ifx\csname urlstyle\endcsname\relax
  \providecommand{\doi}[1]{DOI~\discretionary{}{}{}#1}\else
  \providecommand{\doi}{DOI~\discretionary{}{}{}\begingroup
  \urlstyle{rm}\Url}\fi

\bibitem{Aberdam2018Holistic}
Aberdam, A., {Sulam}, J., {Elad}, M.: {Multi layer sparse coding: the holistic
  way}.
\newblock \emph{To appear} in SIAM Journal on Mathematics of Data Science
  (2018)

\bibitem{aharon2006ksvd}
Aharon, M., Elad, M., Bruckstein, A.: {K-SVD}: An algorithm for designing
  overcomplete dictionaries for sparse representation.
\newblock IEEE Transactions on signal processing \textbf{54}(11), 4311--4322
  (2006)

\bibitem{B17}
Beck, A.: First-order methods in optimization, \emph{MOS-SIAM Series on
  Optimization}, vol.~25.
\newblock Society for Industrial and Applied Mathematics (SIAM),Philadelphia,
  PA (2017)

\bibitem{beck2009fast}
Beck, A., Teboulle, M.: A fast iterative shrinkage-thresholding algorithm for
  linear inverse problems.
\newblock SIAM journal on imaging sciences \textbf{2}(1), 183--202 (2009)

\bibitem{beck2012smoothing}
Beck, A., Teboulle, M.: Smoothing and first order methods: A unified framework.
\newblock SIAM Journal on Optimization \textbf{22}(2), 557--580 (2012)

\bibitem{B83}
Bertsekas, D.P.: Constrained optimization and {L}agrange multiplier methods.
\newblock Computer Science and Applied Mathematics. Academic Press, Inc., New
  York-London (1982)

\bibitem{bioucas2008iterative}
Bioucas-Dias, J.M., Figueiredo, M.A.: An iterative algorithm for linear inverse
  problems with compound regularizers.
\newblock In: Image Processing, 2008. ICIP 2008. 15th IEEE International
  Conference on, pp. 685--688. IEEE (2008)

\bibitem{boyd2011distributed}
Boyd, S., Parikh, N., Chu, E., Peleato, B., Eckstein, J., et~al.: Distributed
  optimization and statistical learning via the alternating direction method of
  multipliers.
\newblock Foundations and Trends{\textregistered} in Machine Learning
  \textbf{3}(1), 1--122 (2011)

\bibitem{candes2000curvelets}
Candes, E.J., Donoho, D.L.: Curvelets: A surprisingly effective nonadaptive
  representation for objects with edges.
\newblock Tech. rep., Stanford Univ Ca Dept of Statistics (2000)

\bibitem{candes2010compressed}
Candes, E.J., Eldar, Y.C., Needell, D., Randall, P.: Compressed sensing with
  coherent and redundant dictionaries.
\newblock arXiv preprint arXiv:1005.2613  (2010)

\bibitem{chartrand2008iterative}
Chartrand, R., Yin, W.: Iteratively reweighted algorithms for compressive
  sensing.
\newblock In: Acoustics, Speech and Signal Processing, 2008. ICASSP 2008. IEEE
  International Conference on, pp. 3869--3872. IEEE (2008)

\bibitem{CT93}
Chen, G., Teboulle, M.: Convergence analysis of a proximal-like minimization
  algorithm using {B}regman functions.
\newblock SIAM J. Optim. \textbf{3}(3), 538--543 (1993).
\newblock \doi{10.1137/0803026}.
\newblock \urlprefix\url{https://doi.org/10.1137/0803026}

\bibitem{combettes2011proximal}
Combettes, P.L., Pesquet, J.C.: Proximal splitting methods in signal
  processing.
\newblock In: Fixed-point algorithms for inverse problems in science and
  engineering, pp. 185--212. Springer (2011)

\bibitem{CW05}
Combettes, P.L., Wajs, V.R.: Signal recovery by proximal forward-backward
  splitting.
\newblock Multiscale Model. Simul. \textbf{4}(4), 1168--1200 (2005)

\bibitem{daubechies2004iterative}
Daubechies, I., Defrise, M., De~Mol, C.: An iterative thresholding algorithm
  for linear inverse problems with a sparsity constraint.
\newblock Communications on pure and applied mathematics \textbf{57}(11),
  1413--1457 (2004)

\bibitem{do2005contourlet}
Do, M.N., Vetterli, M.: The contourlet transform: an efficient directional
  multiresolution image representation.
\newblock IEEE Transactions on image processing \textbf{14}(12), 2091--2106
  (2005)

\bibitem{donoho2003optimally}
Donoho, D.L., Elad, M.: Optimally sparse representation in general
  (nonorthogonal) dictionaries via $\ell_1$ minimization.
\newblock Proceedings of the National Academy of Sciences \textbf{100}(5),
  2197--2202 (2003)

\bibitem{efron2004least}
Efron, B., Hastie, T., Johnstone, I., Tibshirani, R., et~al.: Least angle
  regression.
\newblock The Annals of statistics \textbf{32}(2), 407--499 (2004)

\bibitem{giryes2018tradeoffs}
Giryes, R., Eldar, Y.C., Bronstein, A.M., Sapiro, G.: Tradeoffs between
  convergence speed and reconstruction accuracy in inverse problems.
\newblock IEEE Transactions on Signal Processing \textbf{66}(7), 1676--1690
  (2018)

\bibitem{gregor2010learning}
Gregor, K., LeCun, Y.: Learning fast approximations of sparse coding.
\newblock In: Proceedings of the 27th International Conference on International
  Conference on Machine Learning, pp. 399--406. Omnipress (2010)

\bibitem{haeffele14}
Haeffele, B., Young, E., Vidal, R.: Structured low-rank matrix factorization:
  Optimality, algorithm, and applications to image processing.
\newblock In: E.P. Xing, T.~Jebara (eds.) Proceedings of the 31st International
  Conference on Machine Learning, \emph{Proceedings of Machine Learning
  Research}, vol.~32, pp. 2007--2015. PMLR, Bejing, China (2014)

\bibitem{he2016deep}
He, K., Zhang, X., Ren, S., Sun, J.: Deep residual learning for image
  recognition.
\newblock In: Proceedings of the IEEE conference on computer vision and pattern
  recognition, pp. 770--778 (2016)

\bibitem{henaff2011unsupervised}
Henaff, M., Jarrett, K., Kavukcuoglu, K., LeCun, Y.: Unsupervised learning of
  sparse features for scalable audio classification.
\newblock In: ISMIR, vol.~11, p. 2011. Citeseer (2011)

\bibitem{huang2007unsupervised}
Huang, F.J., Boureau, Y.L., LeCun, Y., et~al.: Unsupervised learning of
  invariant feature hierarchies with applications to object recognition.
\newblock In: Computer Vision and Pattern Recognition, 2007. CVPR'07. IEEE
  Conference on, pp. 1--8. IEEE (2007)

\bibitem{huang2017densely}
Huang, G., Liu, Z., Weinberger, K.Q., van~der Maaten, L.: Densely connected
  convolutional networks.
\newblock In: Proceedings of the IEEE conference on computer vision and pattern
  recognition, vol.~1, p.~3 (2017)

\bibitem{kavukcuoglu2010fast}
Kavukcuoglu, K., Ranzato, M., LeCun, Y.: Fast inference in sparse coding
  algorithms with applications to object recognition.
\newblock arXiv preprint arXiv:1010.3467  (2010)

\bibitem{kutyniok2012shearlets}
Kutyniok, G., Labate, D.: Shearlets: Multiscale analysis for multivariate data.
\newblock Springer Science \& Business Media (2012)

\bibitem{lin2014sparse}
Lin, J., Li, S.: Sparse recovery with coherent tight frames via analysis
  dantzig selector and analysis lasso.
\newblock Applied and Computational Harmonic Analysis \textbf{37}(1), 126--139
  (2014)

\bibitem{mairal2012task}
Mairal, J., Bach, F., Ponce, J.: Task-driven dictionary learning.
\newblock IEEE transactions on pattern analysis and machine intelligence
  \textbf{34}(4), 791--804 (2012)

\bibitem{mairal2010online}
Mairal, J., Bach, F., Ponce, J., Sapiro, G.: Online learning for matrix
  factorization and sparse coding.
\newblock Journal of Machine Learning Research \textbf{11}(Jan), 19--60 (2010)

\bibitem{M65}
Moreau, J.J.: Proximit\'{e} et dualit\'{e} dans un espace hilbertien.
\newblock Bull. Soc. Math. France \textbf{93}, 273--299 (1965)

\bibitem{bruna2017UnderstandingTS}
Moreau, T., Bruna, J.: Understanding trainable sparse coding via matrix
  factorization.
\newblock In: Proceedings of the International Conference on Learning
  Representations (2017)

\bibitem{murdock2018deep}
Murdock, C., Chang, M.F., Lucey, S.: Deep component analysis via alternating
  direction neural networks.
\newblock arXiv preprint arXiv:1803.06407  (2018)

\bibitem{papyan2016convolutional}
Papyan, V., Romano, Y., Elad, M.: Convolutional neural networks analyzed via
  convolutional sparse coding.
\newblock The Journal of Machine Learning Research \textbf{18}(1), 2887--2938
  (2017)

\bibitem{papyan2017working}
Papyan, V., Sulam, J., Elad, M.: Working locally thinking globally: Theoretical
  guarantees for convolutional sparse coding.
\newblock IEEE Transactions on Signal Processing \textbf{65}(21), 5687--5701
  (2017)

\bibitem{sulam2016trainlets}
Sulam, J., Ophir, B., Zibulevsky, M., Elad, M.: Trainlets: Dictionary learning
  in high dimensions.
\newblock IEEE Transactions on Signal Processing \textbf{64}(12), 3180--3193
  (2016)

\bibitem{sulam2017multi}
Sulam, J., Papyan, V., Romano, Y., Elad, M.: Multilayer convolutional sparse
  modeling: Pursuit and dictionary learning.
\newblock IEEE Transactions on Signal Processing \textbf{66}(15), 4090--4104
  (2018)

\bibitem{tibshirani2011regression}
Tibshirani, R.: Regression shrinkage and selection via the lasso: a
  retrospective.
\newblock Journal of the Royal Statistical Society: Series B (Statistical
  Methodology) \textbf{73}(3), 273--282 (2011)

\bibitem{tibshirani2015statistical}
Tibshirani, R., Wainwright, M., Hastie, T.: Statistical learning with sparsity:
  the lasso and generalizations.
\newblock Chapman and Hall/CRC (2015)

\bibitem{tibshirani2011solution}
Tibshirani, R.J.: The solution path of the generalized lasso.
\newblock Stanford University (2011)

\bibitem{tropp2006just}
Tropp, J.A.: Just relax: Convex programming methods for identifying sparse
  signals in noise.
\newblock IEEE transactions on information theory \textbf{52}(3), 1030--1051
  (2006)

\bibitem{wright2010sparse}
Wright, J., Ma, Y., Mairal, J., Sapiro, G., Huang, T.S., Yan, S.: Sparse
  representation for computer vision and pattern recognition.
\newblock Proceedings of the IEEE \textbf{98}(6), 1031--1044 (2010)

\bibitem{zeiler2010deconvolutional}
Zeiler, M.D., Krishnan, D., Taylor, G.W., Fergus, R.: Deconvolutional networks.
\newblock In: Computer Vision and Pattern Recognition (CVPR), 2010 IEEE
  Conference on, pp. 2528--2535. IEEE (2010)

\bibitem{zhang2018ista}
Zhang, J., Ghanem, B.: Ista-net: Interpretable optimization-inspired deep
  network for image compressive sensing.
\newblock In: Proceedings of the IEEE Conference on Computer Vision and Pattern
  Recognition, pp. 1828--1837 (2018)

\end{thebibliography}

\end{document}